\documentclass{article}
\usepackage{iclr2027_conference,times}

\usepackage{amsmath,amsfonts,bm}

\def\eqref#1{equation~\ref{#1}}

\def\1{\bm{1}}

\DeclareMathAlphabet{\mathsfit}{\encodingdefault}{\sfdefault}{m}{sl}
\SetMathAlphabet{\mathsfit}{bold}{\encodingdefault}{\sfdefault}{bx}{n}

\usepackage[dvipsnames]{xcolor}

\RequirePackage{fix-cm}
\usepackage{siunitx}
\usepackage[T1]{fontenc}
\usepackage{times}
\usepackage{amsmath,amssymb,caption,xspace}
\usepackage{tikz}
\usetikzlibrary{arrows.meta,backgrounds,calc}
\DeclareMathSizes{7.5}{7.5}{5}{5}
\DeclareMathSizes{8.5}{8.5}{6}{5}

\definecolor{ink}{HTML}{243444}
\definecolor{tensorblue}{HTML}{356D99}
\definecolor{learnedgold}{HTML}{A96C20}
\definecolor{flowgray}{HTML}{667887}
\definecolor{panelgray}{HTML}{D1D9E0}

\usepackage{amsmath,amssymb,booktabs}
\usepackage{graphicx}
\usepackage{url}
\usepackage{amsmath}
\usepackage{amssymb}
\usepackage{amsthm}
\usepackage{xspace}
\usepackage{multirow}
\usepackage{tikz}
\usetikzlibrary{positioning, arrows.meta, fit, backgrounds, calc}

\newcommand{\methodname}{\textit{DCOCO}\xspace}

\definecolor{citecolor}{HTML}{0071BC}
\definecolor{linkcolor}{HTML}{ED1C24}
\usepackage[pagebackref=true, breaklinks=true, colorlinks,
            citecolor=citecolor, linkcolor=linkcolor, bookmarks=false]{hyperref}

\usepackage{pgfplots}
\pgfplotsset{compat=1.18}
\usepgfplotslibrary{fillbetween}
\definecolor{covkA}{HTML}{86B6EF}
\definecolor{covkB}{HTML}{3987E5}
\definecolor{covkC}{HTML}{1C5CAB}
\definecolor{covkD}{HTML}{0D366B}
\definecolor{verA}{HTML}{2A78D6}
\definecolor{verB}{HTML}{EB6834}
\definecolor{verC}{HTML}{1BAF7A}
\definecolor{chartgrid}{HTML}{E6E5E1}
\definecolor{chartaxis}{HTML}{8A8983}
\pgfplotsset{covstyle/.style={scale only axis, height=3.3cm, xmin=0.5, xmax=15.5, xtick={1,5,10,15}, xlabel={Drafted slot}, grid=major, grid style={line width=0.3pt, draw=chartgrid}, axis line style={draw=chartaxis, line width=0.4pt}, major tick length=0pt, every axis label/.append style={font=\footnotesize}, ticklabel style={font=\scriptsize}, title style={font=\footnotesize, yshift=-4pt}, axis lines*=left, clip=false}}

\theoremstyle{plain}

\providecommand{\todo}[1]{\textcolor{red}{[#1]}}
\providecommand{\tnum}{\textcolor{red}{$\cdot$}}

\providecommand{\pmstd}[2]{#1{\scriptsize\,$\pm$#2}} 
\providecommand{\best}[1]{\textbf{#1}}            

\newcommand{\coconut}{%
  \raisebox{-0.15em}{\includegraphics[height=1.05em]{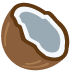}}%
}

\theoremstyle{definition}

\title{\centering Contextual Correction of Parallel Drafts\\ for Speculative Decoding \coconut}

\author{
\makebox[\textwidth][c]{%
\begin{tabular}{cccc}
\textbf{Roy Eisenstadt}$^{1,2}$ &
\textbf{Harel Lidar}$^{1}$ &
\textbf{Tal Gonen}$^{1}$ &
\textbf{Omer Luxembourg}$^{3}$ \\
\quad \quad \textbf{Itamar Zimerman}$^{1}$ &
\textbf{Lior Wolf}$^{2}$ &
\textbf{Eli David}$^{1}$ &
\textbf{Ilan Naiman}$^{1}$
\end{tabular}
}
}

\iclrfinalcopy

\begin{document}

\begingroup
\renewcommand{\thefootnote}{}
\footnotetext{

$^{1}$ Stealth Startup, Tel Aviv \qquad
$^{2}$  Tel Aviv University \qquad
$^{3}$ Ben-Gurion University
}
\addtocounter{footnote}{-1}
\endgroup

\maketitle
\lhead{Preprint}

\begin{abstract}

Speculative decoding accelerates large language model inference by using a lightweight drafter to propose multiple tokens that are subsequently verified in parallel by the target model. Recent diffusion-based drafters make this paradigm particularly attractive: deep attention-based neural backbones provide strong predictions, while drafting multiple future positions in parallel avoids the sequential cost of autoregressive generation. This parallelism, however, comes at a cost. Tokens within a draft block are predicted without access to the realized identities of preceding draft tokens, removing intra-block dependencies that would naturally be available under autoregressive generation. We introduce \methodname (\textbf{D}Flash \textbf{Co}ntextual \textbf{Co}rrector), a lightweight mechanism that restores these dependencies while preserving parallel drafting. Rather than committing to each position's top prediction, \methodname retains multiple candidate tokens and evaluates their contextual compatibility across adjacent draft positions in parallel, enabling contextually informed token selection at low additional cost. Across four verifier backbones and ten benchmarks, \methodname achieves the highest average throughput in every evaluated configuration, spanning greedy and stochastic decoding with and without extended reasoning, with speedups of up to 6.6× over vanilla decoding. Its advantage persists under concurrent serving: \methodname outperforms the strongest competing method by 6–23\% across 4–32 concurrent requests, while other sequential correctors fall below uncorrected DFlash at these loads. These results demonstrate that recovering intra-block dependencies through parallel contextual scoring delivers consistent acceleration across decoding regimes and serving loads.

\centering \textbf{Links:}\hspace{2pt} \href{https://github.com/two-delta/DCOCO/}{Code} (GitHub) $\vert$ \href{https://huggingface.co/collections/TD-Labs/dcoco}{Models} (Hugging Face)

\end{abstract}

\providecommand{\methodname}{\textit{DCOCO}\xspace}

\section{Introduction}
\label{sec:intro}

Speculative decoding (SD) accelerates large language model inference by using a lightweight
drafter to propose multiple future tokens, which are subsequently verified in parallel by
the target model~\citep{leviathan2023fast,chen2023accelerating}. Speculative verification preserves the target model's output distribution, so the benefit of a drafting method depends on the speedup it provides. This speedup depends on both how many draft tokens are accepted and the latency of draft generation. These two factors are often in tension: stronger drafting can increase acceptance, but better drafts may require more computation and therefore more latency.

This acceptance--latency trade-off has shaped the design of speculative drafters. Early SD methods use a smaller autoregressive language model to generate draft tokens sequentially~\citep{leviathan2023fast,chen2023accelerating}. Subsequent approaches such as EAGLE~\citep{li2024eagle,li2024eagle2,li2025eagle3} improve the efficiency and quality of this drafting process by exploiting target-model features, while retaining autoregressive dependencies across draft positions. More recently, parallel diffusion-based drafters such as DFlash~\citep{chen2026dflash} take a different direction, generating multiple future positions within a single block pass. Autoregressive drafters condition on the realized prefix, whereas parallel diffusion drafters reduce latency by predicting the block jointly. This parallelism comes at the cost of weaker intra-block conditioning: each position is
predicted without observing the realized token identities selected at preceding draft
positions. As a result, individually plausible predictions may form a less consistent
continuation and reduce draft acceptance.

Recent methods directly target this limitation. DSpark~\citep{cheng2026dspark}
combines a parallel backbone with a lightweight sequential module, introducing
intra-block dependencies through semi-autoregressive drafting. Domino~\citep{huang2026domino}
instead augments parallel block generation with a lightweight causal correction module
that refines the draft using prefix-dependent information. Both recover acceptance, but both do so position by position. The added module is a
recurrence over the block, so its latency cost sits on the critical path of every drafting
cycle. This raises the central question of our work: \emph{can intra-block
dependencies be restored without reintroducing a sequential module?}

Two observations guide our approach. First, the parallel drafter has usually already suspected the correct token, even when
it did not choose it: on Qwen3-8B the top-16 candidates contain the verifier's token
at $59\%$ of the slots where its top-1 is wrong (\S\ref{sec:prelim}). Correction can
therefore be cast as re-selection within the drafter's existing candidates rather than
generation of new tokens. Re-selection over $K$ candidates per slot reduces to
scoring $K^2$ transitions between adjacent slots, all of which can be computed at once. Second, speculative verification imposes a strong asymmetry across draft positions. Verification accepts a contiguous prefix and stops at the first rejected token; therefore, an improvement at a later position is useful only if all preceding positions are also accepted. Correction should consequently prioritize the quality of the early prefix, rather than optimize an aggregate score over the entire proposed block.

These observations motivate \textbf{D}Flash \textbf{Co}ntextual
\textbf{Co}rrector (\textbf{\methodname{}}), a lightweight contextual corrector for
parallel diffusion drafts. \methodname operates on a single DFlash block pass:
rather than committing immediately to the top prediction at each position,
it retains multiple candidate tokens and uses the drafter's existing hidden
states and logits to score their contextual compatibility across adjacent
positions. All candidate interactions are scored in parallel, without an
additional drafter forward pass. The resulting candidate grid is then decoded using a \emph{prefix-first greedy selection} rule. Rather than maximizing an aggregate score over the entire path, \methodname{} selects candidates from left to right, prioritizing each position before proceeding to the next. This aligns selection with the prefix structure of speculative verification, where tokens after the first rejection provide no benefit. Importantly, only this lightweight selection is sequential, while neural scoring remains parallel. Unlike DSpark and Domino, no learned module runs position by position.

We evaluate \methodname{} across multiple verifier backbones and ten benchmarks spanning mathematics, code, and general-purpose generation. \methodname{} consistently improves end-to-end decoding throughput under both greedy and stochastic decoding, outperforming the strongest evaluated speculative decoding baseline by up to $18\%$ and vanilla autoregressive decoding by up to $6.6\times$. Because its correction is parallel, \methodname{} keeps its advantage as serving load
grows: from 1 to 32 concurrent streams its throughput stays $19$--$25\%$ above DFlash under greedy decoding,
while the per-cycle cost of the sequential correctors grows until both fall below the
uncorrected drafter (\S\ref{sec:exp:concurrency}). Beyond aggregate throughput, we separately analyze draft acceptance and correction latency to characterize where these gains arise. We further compare alternative selection rules over the same learned candidate scores,
isolating the effect of the decoding objective from the quality of the corrector itself. Our contributions are threefold:
\begin{itemize}
    \item We introduce \textbf{\methodname{}}, a lightweight contextual corrector for parallel
    diffusion drafts that scores candidate interactions using information from a single
    drafter pass, without additional drafter forward passes.

    \item We identify the prefix structure of speculative verification as a key
    consideration for correcting parallel drafts, and introduce a prefix-first greedy
    selection rule that aligns candidate selection with this structure.

    \item We establish state-of-the-art performance across multiple
    verifier backbones and diverse benchmarks, with and without extended thinking in both greedy and stochastic decoding
    and show that this advantage persists under concurrent serving load, where the sequential correctors lose theirs.
\end{itemize}

\section{Related Work}

\paragraph{Speculative decoding.} Speculative decoding accelerates autoregressive LLM inference by separating generation into drafting and verification: a lightweight draft model proposes several future tokens, which are then verified in parallel by the target model while preserving the target model's output distribution~\citep{leviathan2023fast,chen2023accelerating}. Early approaches used a smaller autoregressive language model as the drafter, reducing the cost of proposing tokens while retaining autoregressive conditioning across the draft sequence~\citep{leviathan2023fast,chen2023accelerating}. Subsequent work has improved the acceptance--latency trade-off by constructing
lightweight drafters from representations of the target model itself. EAGLE~\citep{li2024eagle} performs autoregression at the feature level, predicting target-model features rather than relying on an independent language-model drafter. EAGLE-2~\citep{li2024eagle2} further introduces context-aware dynamic draft trees, allocating speculative candidates according to the drafter's confidence, while EAGLE-3~\citep{li2025eagle3} replaces feature prediction with direct token prediction
and fuses features from multiple target-model layers. Other approaches explore recurrent drafters conditioned on target-model representations,
such as ReDrafter~\citep{zhang2024redrafter}, or improve the alignment between the
drafter and target representations~\citep{zhang2025hass}.
These methods substantially strengthen lightweight autoregressive drafting, but draft tokens along a candidate path remain sequentially dependent. This has motivated a complementary line of work that moves parallelism from verification into the drafting
stage itself.

\begin{figure}[t]
    \centering
    \includegraphics[width=\linewidth]{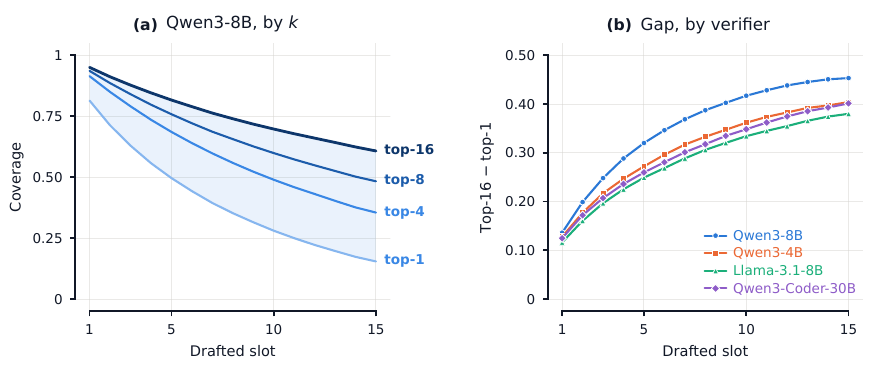}
    \caption{\textbf{Candidate coverage gains grow deeper in the block.}
    Fraction of slots where DFlash's top-$k$ candidates contain the
    verifier's committed token.
    (a) Qwen3-8B; shading marks the top-16--top-1 coverage gap.
    (b) The same gap across verifiers.
    Averages over the 15 slots appear in Tab.~\ref{tab:coverage}.}
    \label{fig:coverage}
\end{figure}

\paragraph{Parallel drafters.} Parallel drafters reduce draft-generation latency by predicting multiple future
positions within a single pass. Medusa~\citep{cai2024medusa} predicts future tokens in parallel using multiple lightweight heads, to which Hydra~\citep{ankner2024hydra} adds sequential dependency in order to increase performance. PARD~\citep{an2026pard} adapts autoregressive drafters to parallel block prediction, while CTC-based drafting~\citep{wen2024speculative} uses a non-autoregressive sequence generation technique to predict the speculated sequence in parallel. Diffusion models provide a natural
foundation for this setting because multiple positions can be denoised simultaneously. DFlash~\citep{chen2026dflash} exploits this property with a lightweight diffusion drafter conditioned on target-model features, generating an entire draft block in one forward pass. However, unlike autoregressive drafting, parallel block prediction does not condition each position on the realized tokens selected at earlier positions, weakening dependencies within the draft and motivating methods that restore this
missing context.

\paragraph{Refining parallel drafts.}

Recent methods address this limitation by augmenting parallel drafts with lightweight causal correction. DSpark~\citep{cheng2026dspark} combines a parallel backbone with a
sequential module that introduces dependencies across draft positions, while Domino~\citep{huang2026domino} applies recurrent causal correction to refine the parallel proposal using prefix-dependent information. 

These approaches highlight the central challenge: recovering the dependencies lost during parallel drafting without sacrificing its latency advantage. \methodname takes a different approach: rather than sequentially refining the draft or iteratively updating the block, it operates on the candidate set produced by a single DFlash pass. \methodname scores all adjacent candidate interactions in one batched operation using the drafter's hidden states and logits, without an additional drafter pass. Only the subsequent candidate selection, reduced to a plain walk on a grid, is sequential. 

\begin{figure*}[!t]
    \centering
    \providecommand{\methodname}{DCOCO}
\definecolor{dcink}{HTML}{26333D}
\definecolor{dcgray}{HTML}{6B7882}
\definecolor{dcblue}{HTML}{527D9C}
\definecolor{dcbluefill}{HTML}{EDF4F8}
\definecolor{dcwarm}{HTML}{B17A42}
\definecolor{dcwarmfill}{HTML}{FCF0E1}
\definecolor{dcgreen}{HTML}{39785B}
\definecolor{dcgreenfill}{HTML}{E8F3EC}
\definecolor{dcline}{HTML}{D6DFE4}
\definecolor{dcbg}{HTML}{FAFCFD}

\begin{minipage}{\linewidth}
\centering
\pgfmathsetlengthmacro{\dcxunit}{\linewidth/139.7}
\begin{tikzpicture}[
 x=\dcxunit,y=1mm,yscale=.775,
 every node/.style={scale=.85},
 font=\sffamily\fontsize{8}{9}\selectfont,text=dcink,
 flow/.style={draw=dcgray,line width=.55pt,-{Latex[length=1.2mm,width=.9mm]},rounded corners=.7mm,shorten >=.3pt},
 wire/.style={draw=dcgray,line width=.55pt,rounded corners=.7mm},
 block/.style={draw=dcwarm,fill=dcwarmfill,rounded corners=2.3pt,line width=.55pt,align=center,inner sep=1.6pt},
 title/.style={anchor=west,font=\sffamily\bfseries\fontsize{8.8}{9.8}\selectfont},
 note/.style={font=\sffamily\fontsize{7}{8}\selectfont,text=dcgray},
 cell/.style={draw=dcgray!80,fill=white,rounded corners=2pt,line width=.65pt,minimum width=23mm,minimum height=6.5mm,inner sep=0pt},
 chosen/.style={cell,draw=dcwarm,fill=dcwarmfill,line width=.85pt},
 draftcell/.style={cell,draw=dcblue,fill=dcbluefill},
 selected/.style={draw=dcwarm,line width=1.15pt,-{Latex[length=1.5mm,width=1.1mm]},shorten >=.3pt},
 draftpath/.style={draw=dcblue,line width=.8pt,dash pattern=on 1.4pt off 1.3pt},
 latent/.style={draw=dcgray!75,line width=.45pt},
 active/.style={draw=dcgray,line width=.6pt}
]
\path[use as bounding box] (0,1) rectangle (139.7,-73);
\node[title] at (1,-3) {(a)\quad Parallel correction};
\node[note,anchor=east] at (138,-3) {\methodname{}: all $(B-1)K^2$ scores in parallel};

\node[note,align=center,font=\sffamily\fontsize{6.8}{7.6}\selectfont] at (10,-10)
 {$x$ + $B$ masks};
\node[block,draw=dcblue,fill=dcbluefill,minimum width=18mm,minimum height=10mm]
 (draft) at (10,-19) {\textbf{DFlash}\\[-1pt]\fontsize{7}{8}\selectfont one pass};
\foreach \x/\i/\one/\two/\three in {
 29/1/These/The/A,43/2/key/keys/locks,57/3/is/are/was}{
 \node[block,draw=dcblue!80,fill=dcbluefill!45,minimum width=12mm,minimum height=17mm,
 font=\sffamily\fontsize{7}{7.8}\selectfont] (slot\i) at (\x,-18.5)
 {\textbf{Slot \i}\\[1pt]\texttt{\one}\\[-1pt]\texttt{\two}\\[-1pt]\texttt{\three}\\[1pt]
  \textcolor{dcblue}{$h_{\i},\ell_{\i}$}};
}
\draw[wire] (draft.east) -- (20.5,-19) -- (20.5,-7.5) -- (57,-7.5);
\foreach \x/\i in {29/1,43/2,57/3}{\draw[flow] (\x,-7.5) -- (slot\i.north);}
\node[text=dcblue,font=\fontsize{8}{9}\selectfont,inner sep=0pt] at (66,-20) {$\cdots$};
\foreach \x/\i in {29/1,43/2,57/3}{\draw[flow] (slot\i.south) -- (\x,-29.5);}
\draw[wire] (29,-29.5) -- (67,-29.5) -- (67,-13);

\begin{scope}[on background layer]
 \filldraw[draw=dcwarm!60,fill=dcwarmfill!20,rounded corners=3pt,line width=.5pt]
 (69.5,-5.8) rectangle (139,-35);
\end{scope}
\node[block,minimum width=24mm,minimum height=12mm] (cand) at (83,-13)
 {\textbf{Candidate}\\[-1pt]\textbf{representations}\\[1pt]
 \fontsize{6.8}{7.6}\selectfont Per top-$K$ candidate};
\node[block,minimum width=24mm,minimum height=10mm] (pair) at (83,-26.1)
 {\textbf{Pair state}\\[1pt]
 \fontsize{6.8}{7.6}\selectfont Per adjacent slot pair};
\draw[flow] (67,-13) -- (cand.west);
\draw[flow] (67,-26.1) -- (pair.west);
\node[block,minimum width=17mm,minimum height=12mm] (mlp) at (106,-18)
 {\textbf{Shared MLP}\\[-1pt]\fontsize{7}{8}\selectfont two layers\\[1pt]
 \fontsize{6.8}{7.6}\selectfont Per pair $(a,b)$};
\draw[flow] (cand.east) -- ++(1.3,0) |- ([yshift=2.5mm]mlp.west);
\draw[flow] (pair.east) -- ++(1.3,0) |- ([yshift=-2.5mm]mlp.west);
\node[circle,draw=dcwarm,fill=dcwarmfill,line width=.6pt,minimum size=4.5mm,inner sep=0pt]
 (plus) at (120,-18) {$+$};
\draw[flow] (mlp.east) -- (plus.west);
\node[note,text=dcink] at (120,-12.5) {$\Delta_\theta$};
\node[text=dcwarm,anchor=west,inner sep=0pt,font=\fontsize{7.8}{8.8}\selectfont]
 (scores) at (126,-18) {$S_p(a,b)$};
\draw[flow,draw=dcwarm] (plus.east) -- (scores.west);
\draw[wire,draw=dcblue] (57,-29.5) -- (57,-33) -- (120,-33);
\draw[flow,draw=dcblue] (106,-33) -- (mlp.south);
\draw[flow,draw=dcblue] (120,-33) -- (plus.south);
\node[note,text=dcblue,anchor=west,fill=dcwarmfill!20,inner sep=.3pt]
 at (107,-29) {logits};
\node[note,text=dcblue,anchor=west] at (121,-29) {$\ell_{p+1}(b)$};

\node[title] at (1,-39) {(b)\quad Prefix-first selection};
\node[note,anchor=east] at (138,-39) {Rows follow draft rank};
\foreach \x/\i in {15/1,43/2,71/3,99/4}{
 \foreach \y/\k in {-50/1,-59/2,-68/3}{\coordinate (n\i\k) at (\x,\y);}
}
\node[note] at (15,-44.5) {Slot 1: $\ell_1(k)$};
\node[note] at (43,-44.5) {Slot 2: $S_1(k_1,k)$};
\node[note] at (71,-44.5) {Slot 3: $S_2(k_2,k)$};
\node[note] at (99,-44.5) {Slot 4: $S_3(k_3,k)$};

\foreach \name/\word/\value/\style in {
 11/These/4.2/chosen,12/The/3.6/cell,13/A/2.8/cell,
 21/key/0.4/draftcell,22/keys/2.7/chosen,23/locks/1.1/cell,
 31/is/0.2/draftcell,32/are/3.1/chosen,33/was/0.8/cell,
 41/missing/2.8/chosen,42/here/1.6/cell,43/broken/0.5/cell}{
 \node[\style] (c\name) at (n\name) {};
 \node[anchor=west,font=\ttfamily\fontsize{7.8}{8.8}\selectfont,inner sep=0pt]
 at ([xshift=1.4mm]c\name.west) {\word};
 \node[note,anchor=east,inner sep=0pt] at ([xshift=-1.4mm]c\name.east) {\value};
}
\begin{scope}[on background layer]
 \foreach \left/\right in {1/2,2/3,3/4}{\foreach \a in {1,2,3}{\foreach \b in {1,2,3}{
  \draw[latent] (c\left\a.east) -- (c\right\b.west);
 }}}
 \foreach \from/\next in {11/2,22/3,32/4}{\foreach \k in {1,2,3}{
  \draw[active] (c\from.east) -- (c\next\k.west);
 }}
 \foreach \from/\to in {11/21,21/31,31/41}{
  \draw[draftpath] (c\from.east) -- (c\to.west);
 }
 \foreach \from/\to in {11/22,22/32,32/41}{
  \draw[selected] (c\from.east) -- (c\to.west);
 }
\end{scope}
\node[note,align=center] at (126,-48) {Select from the\\chosen predecessor};
\draw[draftpath] (114,-57) -- (119,-57);
\node[note,anchor=west,text=dcblue,inner sep=1pt] at (121,-57) {Draft top-1};
\draw[selected] (114,-65) -- (119,-65);
\node[note,anchor=west,text=dcwarm,inner sep=1pt] at (121,-65) {DCOCO};
\end{tikzpicture}
\ifdefined\DCOCOdiagramonly\else
\captionsetup{font=footnotesize,labelfont=bf,skip=3pt}
\captionof{figure}{\textbf{\methodname{}: parallel scoring, prefix-first selection.}
(a) One DFlash pass supplies slot states, candidate IDs, and logits. Candidate and pair
representations feed a shared MLP, computing all residual scores in parallel:
$S_p(a,b)=\ell_{p+1}(b)+\Delta_\theta$.
(b) Starting from the first-slot draft argmax, selection follows the best transition from each
chosen predecessor. Rows retain draft rank; scores in slots 2--4 condition on the selected
predecessor. Tokens and scores are illustrative.}
\label{fig:dcoco}
\fi
\end{minipage}
\end{figure*}
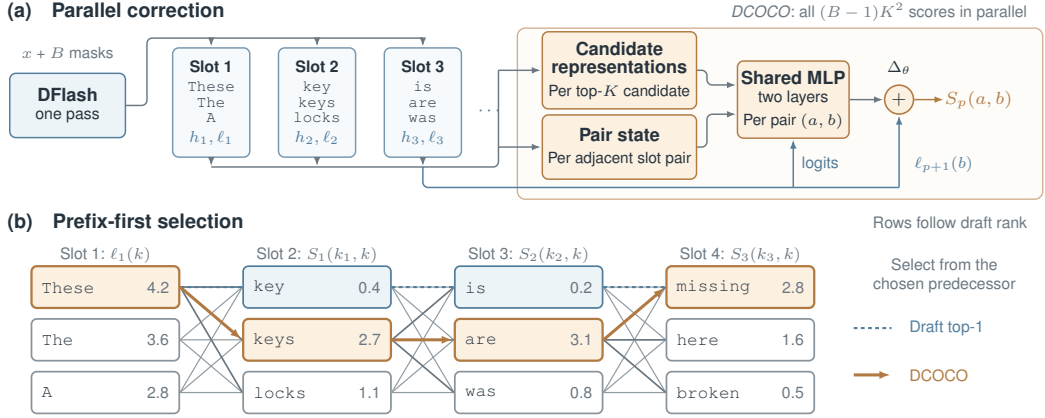

\section{Parallel drafts already contain the right token}
\label{sec:prelim}

A parallel drafter predicts all block positions without
conditioning on the tokens selected at earlier positions.
This can produce incoherent combinations of individually
plausible predictions, a known limitation of non-autoregressive
and discrete-diffusion language modeling
\citep{gu2018nonautoregressive,lee2026flm}.
For example, a drafter assigning probability to both ``New York''
and ``San Diego'' may emit ``New Diego''.
Such errors raise the question of whether the correct token
remains among the drafter's leading candidates. To test whether correct alternatives remain available for
re-selection, we measure \emph{top-$k$ coverage}: the fraction of drafted
slots whose $k$ highest-scoring candidates contain the token
eventually committed by the verifier. We record DFlash's
candidates during speculative decoding at $T=0$ and compare
them with the realized continuation. We observe two regularities.

Fig.~\ref{fig:coverage} shows that on Qwen3-8B, top-16 coverage stays
above $0.60$ at the last slot while top-1 falls below $0.16$, and the
top-16$\,-\,$top-1 gap grows from $0.12$--$0.14$ at the first slot to
$0.38$--$0.45$ at the last on \emph{every} verifier
(averages in Tab.~\ref{tab:coverage}). The drafter does not lack the
right tokens; it fails to \emph{select} them, increasingly so deeper
in the block. That headroom is conditional, however: under prefix
acceptance a correction at slot $p$ counts only if every earlier slot
is accepted, hence our prefix-first rule, selecting candidates left
to right, each conditioned on the choice just made.

\section{Method}
\label{sec:method}

We consider speculative decoding with a parallel block drafter: DFlash proposes a block of $B$ tokens in a single forward pass, and a verifier accepts the longest correct prefix of the proposal. Our proposed corrector, \methodname, works in two steps. It first reassigns scores to all drafter's per-slot candidates based on stronger contextual information given the \emph{realized} tokens and second, it re-selects the proposed block based on the scoring. Figure~\ref{fig:dcoco} gives an overview of the architecture and the
flow from drafted block to corrected proposal.

\subsection{Block drafting and prefix acceptance}
The verifier scores the proposed block $\hat y_{1:B}$ in a single pass
and accepts a prefix of it, judging slots left to right, then commits
one further token of its own. Under greedy decoding, slot $j$ is
accepted iff $\hat y_j$ matches the verifier's greedy choice
$\arg\max_v q(v \mid x, \hat y_{<j})$; under stochastic decoding it is
accepted by the standard speculative-sampling rule in \cite{leviathan2023fast} which accepts $\hat y_j$ with probability $\min\bigl\{1,\, q(\hat y_j \mid x, \hat y_{<j}) / p(\hat y_j \mid x,
\hat y_{<j})\bigr\}$ and otherwise resamples from the residual. \methodname selects proposals deterministically in both decoding regimes, so its proposal distribution is a point mass at each selected token; the resulting lossless stochastic verification rule is detailed in Appendix~\ref{app:deterministic-proposals}. In
both regimes, verification halts at the first rejection. Writing
$a_j \in \{0,1\}$ for slot $j$'s acceptance, the accepted length is
\begin{equation}
\label{eq:accepted}
A(\hat y) \;=\; \max\bigl\{\, m :\, a_1 = \dots = a_m = 1 \,\bigr\}\,,
\end{equation}
and each round commits $A(\hat y)+1$ tokens. Throughput is thus
governed by the expected accepted length together with the wall-clock
cost of a round. The one structural fact that drives every design choice below is that verification halts at the first rejection: a corrected token at
slot~$5$ standing behind a rejection at slot~$1$ is never even
examined. Expected accepted length depends on the probability that each successive prefix survives verification. This motivates prioritizing early positions.

\subsection{A learned corrector over the block grid}
\label{sec:method:corrector}

At each slot $i \in \{1,\ldots,B\}$, DFlash produces a hidden
state $h_i \in \mathbb{R}^d$ and retains its top-$K$ candidate
tokens. Let $c_{ik}$ denote the token id of the $k$-th retained
candidate and $\ell_i(k)$ its draft logit. These candidates form
a dense first-order grid with $B \times K$ nodes, in which every
candidate at slot $p$ connects to every candidate at slot $p+1$.
The \methodname{} corrector assigns each transition a score
\begin{equation}
\label{eq:edge}
S_p(a,b)
=
\underbrace{\ell_{p+1}(b)}_{\text{draft unary logit}}
+
\underbrace{
\Delta_\theta\bigl(
s_p,\; z_{pa},\; z_{p+1,b},\;
\ell_p(a),\; \ell_{p+1}(b)
\bigr)
}_{\text{learned correction}}\,,
\end{equation}
where $a,b \in \{1,\ldots,K\}$ index the candidates at the two
slots, $z_{ik}$ is a candidate representation
(Eq.~\ref{eq:zik}), $s_p$ is a candidate-independent pair state
(Eq.~\ref{eq:pair}), and $\Delta_\theta$ is a shared two-layer
MLP. The corrector uses only the hidden states, candidate ids,
and logits available from the single DFlash pass.
 
\paragraph{Candidate representation.}
Each candidate representation combines the candidate's token
embedding with its slot's hidden state. Using the frozen
token-embedding table $E$ and layer normalization $\mathrm{LN}$,
we compute
\vspace{2mm}
\begin{align}
b_{ik}
&= \mathrm{LN}\bigl(W_{\mathrm{cand}} E[c_{ik}]\bigr)\,,
&
g_i
&= \mathrm{LN}\bigl(W_{\mathrm{ctx}} h_i\bigr)\,,
\label{eq:operands}\\
z_{ik}
&= \mathrm{LN}_{\gamma,\beta}\bigl(
W_{\mathrm{fuse}} [\,b_{ik}\,;\,g_i\,]
\bigr)\,,
\label{eq:zik}
\end{align}
\vspace{2mm}
where $b_{ik}$ is the projected and normalized candidate
embedding, $g_i$ is the projected and normalized slot context
shared by all candidates at slot $i$, and $z_{ik}$ is their fused
representation. Here, $[\,\cdot\,;\,\cdot\,]$ denotes concatenation,
$W_{\mathrm{cand}},W_{\mathrm{ctx}} \in \mathbb{R}^{r \times d}$,
and $W_{\mathrm{fuse}} \in \mathbb{R}^{r \times 2r}$, with
corrector width $r \ll d$. The token embeddings and hidden states differ in scale by more
than two orders of magnitude. We therefore use non-affine layer
normalization for both projections in Eq.~\ref{eq:operands}.
The output normalization in Eq.~\ref{eq:zik} includes learnable
scale and bias parameters $\gamma$ and $\beta$.

\begin{figure}[t]
    \centering
    \includegraphics[width=\linewidth]{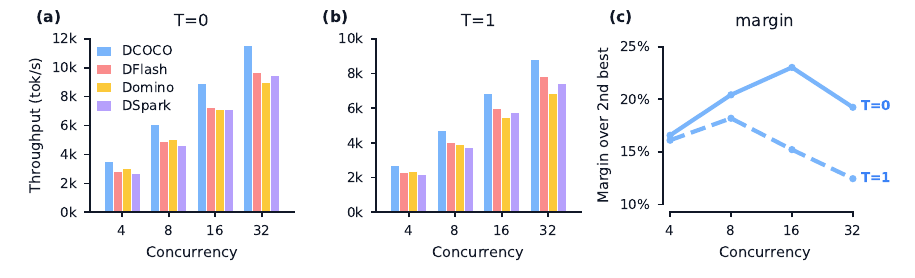}
    \caption{\textbf{At high concurrency, \methodname achieves the highest throughput, while Domino and DSpark fall below DFlash.}
    Qwen3-8B throughput at (a) $T=0$ and (b) $T=1$.
    (c) \methodname's relative throughput gain over the strongest
    baseline at each concurrency.}
    \label{fig:q8b_conc}
\end{figure}

\paragraph{Pair state.}
For each pair of adjacent slots, we compute a shared state from
their hidden states:
\begin{equation}
\label{eq:pair}
s_p
=
W_{\mathrm{pair}} [\,h_p\,;\,h_{p+1}\,]\,,
\qquad
W_{\mathrm{pair}} \in \mathbb{R}^{r \times 2d}\,.
\end{equation}
Because $s_p$ depends only on the slots, it is computed once and
reused across all $K^2$ candidate transitions between them.

\paragraph{Transition score.}
The correction $\Delta_\theta$ in Eq.~\ref{eq:edge} is computed
by a two-layer MLP with a GELU activation between the layers and
a scalar output. Its input is the concatenation
$[\,s_p\,;\,z_{pa}\,;\,z_{p+1,b}\,;\,\ell_p(a)\,;\,\ell_{p+1}(b)\,]$.
The successor's draft logit provides the base score, while
$\Delta_\theta$ learns an additive correction conditioned on
both candidates and their slot contexts. The MLP's output layer
is zero-initialized, so initially $\Delta_\theta \equiv 0$ and
every transition score reduces to the successor's draft logit,
$S_p(a,b)=\ell_{p+1}(b)$. Prefix-first greedy selection therefore
reproduces the drafter's original proposal at initialization. Finally, all projection and MLP parameters are shared across slots and
transitions, and all $(B-1)K^2$ transition scores are computed
in one batched operation before path selection begins.

\paragraph{Training.}
\label{sec:method:training}
 
The drafter and corrector are optimized together using the sum of the original drafter loss and the correction loss. The token-embedding table remains frozen throughout training. For each transition $p$, the label $(a^{\star}_p, b^{\star}_p)$ is the pair of verifier-argmax tokens at the two slots, computed over the teacher-forced context and located within the slots' top-$K$ sets, with mask $m_p \in \{0,1\}$ zeroing transitions where either token is absent. The corrector minimizes an exponentially down-weighted cross-entropy over the $K^2$ jointly softmax-normalized transition
scores,
\begin{equation}
\label{eq:loss}
\mathcal{L}_{\mathrm{corr}}
\;\propto\;
-\sum_{p=1}^{B-1} e^{-(p-1)/\rho}\; m_p\,
\log \frac{\exp S_p(a^{\star}_p, b^{\star}_p)}
          {\sum_{a,b=1}^{K} \exp S_p(a,b)}\,,
\qquad \rho = 4\,.
\end{equation}
This softmax exists only in training; at inference $S_p$ is used
unnormalized. The complete loss term along with further information about the training procedure are detailed in App.~\ref{app:training}. 

\subsection{Prefix-first path selection}
\label{sec:method:greedy}
Given the scored grid, one possible decoding strategy returns the path maximizing the sum of transition scores. This sum does not explicitly account for prefix survival: a higher score at a later position can compensate for a lower score at an earlier position, even though later tokens contribute only if the preceding prefix is accepted. Motivated by this asymmetry, we use a simple prefix-first greedy rule. The first slot keeps the drafter’s top candidate, and each subsequent slot selects the highest-scoring successor of the chosen predecessor,
\begin{equation}
\label{eq:greedy}
k_1 = \arg\max_k\, \ell_1(k)\,,
\qquad
k_{p+1} = \arg\max_b\, S_p(k_p, b)
\quad \text{for } p = 1, \dots, B{-}1\,.
\end{equation}
This rule makes each selection using the transition scores conditioned on the chosen predecessor, without lookahead over future transitions. Its main advantage is computational: the walk costs $(O(BK))$ on top of the $(O(BK^2))$ batched scoring pass. Section 5.3 shows that it achieves acceptance comparable to geometrically discounted global selection at higher throughput. With $(\Delta_\theta \equiv 0)$, it reduces to the drafter’s own argmax at every slot, preserving the step-0 equivalence noted above.


\providecommand{\todo}[1]{\textcolor{red}{[#1]}}
\providecommand{\tnum}{\textcolor{red}{$\cdot$}}
\providecommand{\tspd}{\textcolor{red}{$\cdot$\,{\scriptsize($\cdot\times$)}}}

\providecommand{\methodname}{\textit{DCOCO}\xspace}
\providecommand{\todo}[1]{\textcolor{red}{[#1]}} 
\providecommand{\tnum}{\textcolor{red}{$\cdot$}}  
\providecommand{\pmstd}[2]{#1{\scriptsize\,$\pm$#2}} 
\providecommand{\best}[1]{\textbf{#1}}           
\providecommand{\btau}[1]{\underline{#1}}         

\section{Experiments}
\label{sec:experiments}
\vspace{-3mm}
\paragraph{Setup.}
We use the vLLM framework~\citep{kwon2023vllm} to evaluate \methodname{} against vanilla decoding, DFlash, DSpark, and Domino for four verifier
backbones: Qwen3-(4B/8B/Coder-30B-A3B) and Llama-3.1-8B in production-like settings. 
Evaluation covers ten benchmarks
spanning mathematics, code, and general-purpose generation, under greedy
($T=0$) and stochastic ($T=1$) decoding, with and without extended
thinking where supported. All drafters use block size 16 and are trained
on a common corpus of approximately 800K prompts, with continuations
generated by their respective verifiers. \methodname{} starts from a
six-epoch DFlash checkpoint and trains for six additional epochs with
the corrector; training-budget comparisons are reported in
App.~\ref{app:budget}. Training and implementation details appear in
App.~\ref{app:training}, and evaluation and serving settings in
App.~\ref{app:eval}.

\vspace{-3mm}
\paragraph{Baselines.}
Beyond improvement achieved over vanilla decoding we compare against the uncorrected DFlash draft, DSpark, and Domino. Each is retrained by us on the same corpus with the same five-layer drafter backbone
reading the same verifier layers, and served with the same stack, hardware, and generation limits.
Domino uses its published configuration (causal-encoder width 1024, correction rank 256) and DSpark
a rank-256 Markov head with its confidence head.

\vspace{-3mm}
\paragraph{Metrics.}
Our primary metric is end-to-end decoding throughput in
\textbf{tokens per second} (TPS), measured on a single B200 GPU per system at concurrency~1 unless stated
otherwise; alongside each system's TPS we report its speedup
over vanilla autoregressive decoding with the verifier alone
($\mathrm{TPS}/\mathrm{TPS}_{\text{vanilla}}$). TPS is the quantity a deployment actually
experiences, and it is the appropriate one for a method that adds compute
to the decode loop: a corrector can raise the accepted length while
losing wall-clock time to its own scoring cost, and only TPS exposes
that trade. Every design constraint in \S\ref{sec:method}, the
small corrector width, the single batched scoring pass, the $O(BK)$
walk, exists in service of this metric. We additionally report the mean accepted length $\tau =
\mathbb{E}[A(\hat y)] + 1$ of Eq.~\ref{eq:accepted} as the
\emph{mechanism} metric: it isolates proposal quality from serving
overheads and explains \emph{where} a TPS change comes from ($\tau$ up
$\Rightarrow$ fewer verifier rounds per token). 

\vspace{-2mm}
\subsection{Main results}
\label{sec:exp:main}
\paragraph{Conditions.}
Tab.~\ref{tab:main-q8b} reports per-suite results for Qwen3-8B with thinking off, under
greedy decoding ($T{=}0$) and sampling ($T{=}1$); Tab.~\ref{tab:aggregate} summarizes all four
verifiers, with and without extended thinking, by their ten-suite averages. All numbers are output tokens/s at concurrency~1, with the system served on a single B200 GPU, generating at most
2048 tokens with thinking off and 4096 with thinking on, over the ten
suites above. $T{=}0$ entries are single deterministic runs; $T{=}1$
entries are mean{\scriptsize$\pm$}std over \emph{three} seeds. Bold marks the best speculative method per row and
temperature; vanilla decoding is the
speedup denominator. Per-suite tables for every other configuration, the accepted-length companion of every
throughput table, and evaluation details are in App.~\ref{app:full-results}
and~\ref{app:eval}.

\begin{table}[t]
\centering
\small
\begin{minipage}{\linewidth}
\centering
\textbf{(a)} Greedy decoding, $T{=}0$ (single deterministic run)\\[3pt]
\resizebox{\linewidth}{!}{%
\begin{tabular}{llccccc}
\toprule
Domain & Dataset & No-spec & DFlash & Domino & DSpark & \methodname \\
\midrule
\multirow{3}{*}{Math}
 & MATH-500       & 265.2 & 1293.8 (4.88$\times$) & 1412.3 (5.33$\times$) & 1257.9 (4.74$\times$) & \best{1625.2 (6.13$\times$)} \\
 & AIME-25        & 263.7 & 1133.7 (4.30$\times$) & 1279.9 (4.85$\times$) & 1188.4 (4.51$\times$) & \best{1436.8 (5.45$\times$)} \\
 & GSM8K          & 267.0 & 936.9 (3.51$\times$) & 1018.1 (3.81$\times$) & 921.6 (3.45$\times$) & \best{1175.8 (4.40$\times$)} \\
\midrule
\multirow{3}{*}{Code}
 & HumanEval      & 266.6 & 873.9 (3.28$\times$) & 950.5 (3.57$\times$) & 857.5 (3.22$\times$) & \best{1075.6 (4.04$\times$)} \\
 & MBPP           & 267.0 & 800.9 (3.00$\times$) & 891.7 (3.34$\times$) & 778.2 (2.92$\times$) & \best{979.1 (3.67$\times$)} \\
 & LiveCodeBench  & 264.2 & 1080.1 (4.09$\times$) & 1161.6 (4.40$\times$) & 1084.7 (4.11$\times$) & \best{1328.5 (5.03$\times$)} \\
\midrule
\multirow{4}{*}{General}
 & MT-Bench       & 266.0 & 434.1 (1.63$\times$) & 425.6 (1.60$\times$) & 386.5 (1.45$\times$) & \best{494.7 (1.86$\times$)} \\
 & Arena-Hard-v2  & 264.1 & 423.7 (1.60$\times$) & 390.1 (1.48$\times$) & 386.8 (1.46$\times$) & \best{486.9 (1.84$\times$)} \\
 & Alpaca         & 265.9 & 428.8 (1.61$\times$) & 436.4 (1.64$\times$) & 392.9 (1.48$\times$) & \best{496.1 (1.87$\times$)} \\
 & SpeedBench      & 266.5 & 425.3 (1.60$\times$) & 426.8 (1.60$\times$) & 401.7 (1.51$\times$) & \best{498.4 (1.87$\times$)} \\
\midrule
\multicolumn{2}{l}{Average} & 265.6 & 783.1 (2.95$\times$) & 839.3 (3.16$\times$) & 765.6 (2.88$\times$) & \best{959.7 (3.62$\times$)} \\
\bottomrule
\end{tabular}%
}
\end{minipage}\\[10pt]
\begin{minipage}{\linewidth}
\centering
\textbf{(b)} Sampling, $T{=}1$ (mean{\scriptsize$\pm$}std over seeds 42/43/44)\\[3pt]
\resizebox{\linewidth}{!}{%
\begin{tabular}{llccccc}
\toprule
Domain & Dataset & No-spec & DFlash & Domino & DSpark & \methodname \\
\midrule
\multirow{3}{*}{Math}
 & MATH-500       & \pmstd{264.5}{0.1} & \pmstd{943.9}{37.0} (3.57$\times$) & \pmstd{1013.7}{25.3} (3.83$\times$) & \pmstd{956.5}{33.2} (3.62$\times$) & \best{\pmstd{1145.5}{15.5} (4.33$\times$)} \\
 & AIME-25        & \pmstd{263.6}{0.0} & \pmstd{767.5}{25.5} (2.91$\times$) & \pmstd{781.2}{74.1} (2.96$\times$) & \pmstd{698.5}{14.3} (2.65$\times$) & \best{\pmstd{828.7}{68.3} (3.14$\times$)} \\
 & GSM8K          & \pmstd{266.4}{0.0} & \pmstd{852.8}{16.0} (3.20$\times$) & \pmstd{920.2}{5.6} (3.45$\times$) & \pmstd{834.5}{13.9} (3.13$\times$) & \best{\pmstd{1081.3}{28.5} (4.06$\times$)} \\
\midrule
\multirow{3}{*}{Code}
 & HumanEval      & \pmstd{266.4}{0.0} & \pmstd{770.4}{2.7} (2.89$\times$) & \pmstd{813.7}{18.7} (3.06$\times$) & \pmstd{755.0}{6.1} (2.83$\times$) & \best{\pmstd{910.0}{13.9} (3.42$\times$)} \\
 & MBPP           & \pmstd{266.6}{0.3} & \pmstd{703.1}{4.5} (2.64$\times$) & \pmstd{756.9}{24.6} (2.84$\times$) & \pmstd{678.7}{12.3} (2.55$\times$) & \best{\pmstd{848.3}{6.1} (3.18$\times$)} \\
 & LiveCodeBench  & \pmstd{265.3}{0.3} & \pmstd{874.8}{36.1} (3.30$\times$) & \pmstd{861.8}{23.2} (3.25$\times$) & \pmstd{889.7}{4.4} (3.35$\times$) & \best{\pmstd{1013.7}{62.6} (3.82$\times$)} \\
\midrule
\multirow{4}{*}{General}
 & MT-Bench       & \pmstd{265.3}{0.0} & \pmstd{401.9}{5.1} (1.52$\times$) & \pmstd{402.4}{7.9} (1.52$\times$) & \pmstd{369.4}{8.9} (1.39$\times$) & \best{\pmstd{450.8}{8.0} (1.70$\times$)} \\
 & Arena-Hard-v2  & \pmstd{264.2}{0.1} & \pmstd{379.4}{1.2} (1.44$\times$) & \pmstd{366.6}{11.4} (1.39$\times$) & \pmstd{350.2}{9.0} (1.33$\times$) & \best{\pmstd{436.8}{6.5} (1.65$\times$)} \\
 & Alpaca         & \pmstd{265.5}{0.0} & \pmstd{402.1}{2.7} (1.51$\times$) & \pmstd{393.9}{8.7} (1.48$\times$) & \pmstd{377.4}{1.4} (1.42$\times$) & \best{\pmstd{447.5}{7.0} (1.69$\times$)} \\
 & SpeedBench      & \pmstd{266.2}{0.4} & \pmstd{401.9}{7.9} (1.51$\times$) & \pmstd{398.5}{6.9} (1.50$\times$) & \pmstd{376.7}{2.9} (1.42$\times$) & \best{\pmstd{459.5}{14.9} (1.73$\times$)} \\
\midrule
\multicolumn{2}{l}{Average} & 265.4 & 649.8 (2.45$\times$) & 670.9 (2.53$\times$) & 628.7 (2.37$\times$) & \best{762.2 (2.87$\times$)} \\
\bottomrule
\end{tabular}%
}
\end{minipage}
\caption{\textbf{Throughput (output tokens/s, per-stream) and speedup
over no-spec decoding, Qwen3-8B, thinking off}, concurrency~1, max
2048 generated tokens. Each speculative cell reports TPS with its
speedup over the no-spec column in parentheses. Bold: best
method per row.}
\label{tab:main-q8b}
\end{table}
\vspace{-4mm}
\begin{table}[t]
\centering
\fontsize{7.5}{8.5}\selectfont
\setlength{\tabcolsep}{2pt}

\begin{tabular*}{\linewidth}{@{\extracolsep{\fill}}llcccccccc@{}}
\toprule
& & \multicolumn{2}{c}{DFlash}
  & \multicolumn{2}{c}{Domino}
  & \multicolumn{2}{c}{DSpark}
  & \multicolumn{2}{c}{\methodname} \\
\cmidrule(lr){3-4}\cmidrule(lr){5-6}
\cmidrule(lr){7-8}\cmidrule(lr){9-10}
Verifier & Think.
 & $T{=}0$ & $T{=}1$
 & $T{=}0$ & $T{=}1$
 & $T{=}0$ & $T{=}1$
 & $T{=}0$ & $T{=}1$ \\
\midrule
\multirow{2}{*}{Qwen3-4B}
 & off & 2.98$\times$ & 2.52$\times$ & 3.27$\times$ & 2.71$\times$ & 2.66$\times$ & 2.28$\times$ & \best{3.62$\times$} & \best{2.95$\times$} \\
 & on  & 2.08$\times$ & 1.87$\times$ & 2.17$\times$ & 1.93$\times$ & 1.85$\times$ & 1.69$\times$ & \best{2.38$\times$} & \best{2.12$\times$} \\
\midrule
\multirow{2}{*}{Qwen3-8B}
 & off & 2.95$\times$ & 2.45$\times$ & 3.16$\times$ & 2.53$\times$ & 2.88$\times$ & 2.37$\times$ & \best{3.62$\times$} & \best{2.87$\times$} \\
 & on  & 1.95$\times$ & 1.78$\times$ & 1.97$\times$ & 1.79$\times$ & 1.83$\times$ & 1.68$\times$ & \best{2.26$\times$} & \best{2.03$\times$} \\
\midrule
Qwen3-Coder-30B-A3B
 & off & 1.89$\times$ & 1.67$\times$ & 1.94$\times$ & 1.67$\times$ & 1.84$\times$ & 1.62$\times$ & \best{2.31$\times$} & \best{1.98$\times$} \\
\midrule
Llama-3.1-8B
 & off & 3.03$\times$ & 1.75$\times$ & 3.24$\times$ & 1.79$\times$ & 3.01$\times$ & 1.65$\times$ & \best{3.75$\times$} & \best{1.93$\times$} \\
\bottomrule
\end{tabular*}

\caption{\textbf{All configurations at a glance: average speedup
over vanilla decoding}, unweighted mean over the ten evaluation
suites. Bold: best per row and temperature.
Qwen3-Coder-30B-A3B has no thinking mode.
Per-dataset throughput and accepted length:
App.~\ref{app:full-results}.}
\label{tab:aggregate}
\end{table}

\paragraph{Discussion.}
On Qwen3-8B without thinking, \methodname improves average accepted length over Domino by 8.1\% at T=0 and 7.5\% at T=1, while improving request throughput by 14.3\% and 13.6\%. Accepted length measures proposal quality, whereas request throughput also includes prefill and serving overhead. 

Tab.~\ref{tab:aggregate} shows the same
ordering across verifiers and thinking modes: \methodname has the best average speedup in all twelve cells, by $8$--$19\%$ over the strongest baseline in each; it also has the best average accepted
length in eleven of the twelve (App.~\ref{app:full-results}). On the Qwen3 verifiers the margin under sampling stays within two points of
the greedy margin; on Llama-3.1-8B it halves ($+16\%$ at $T{=}0$, $+8\%$ at $T{=}1$). 

Throughput is the number of tokens committed per verification cycle
divided by the time a cycle takes, so a method can gain on either axis,
and \methodname gains on both. On Qwen3-8B it attains the highest $\tau$ in every cell of both configurations (Tab.~\ref{tab:tau-q8b}
and~\ref{tab:tau-q8b-think}). Relative to the other correction methods,
\methodname both accepts longer prefixes and runs shorter cycles: against the strongest baseline, $7$--$8\%$ longer accepted prefixes and $6\%$ shorter cycles multiply to the $13$--$15\%$ throughput gain. Relative to the uncorrected drafter, it pays $3$--$4\%$ per cycle for $18$--$27\%$ longer accepted prefixes (App.~\ref{app:eval}). Absolute speedups follow the drafter's base acceptance: with thinking
off, math and code reach $3.7$--$6.1\times$ while open-ended chat stays
at $1.8$--$1.9\times$
($2.3$--$2.8\times$ and $1.7$--$2.1\times$ at $T{=}0$ in extended thinking).

App.~\ref{app:budget} ablates the corrector's two input pathways and
its training: removing the pair state costs $7.3\%$/$5.6\%$ of
accepted length ($T{=}0$/$T{=}1$) and removing the candidate
representations $4.8\%$/$3.7\%$, uniformly across the suites, while
freezing the drafter, the corrector as a pure drop-in, still
improves DFlash by $4.4\%$, with joint training worth a further
$\approx\!15\%$.



\subsection{Scaling with concurrency}
\label{sec:exp:concurrency}

\begin{figure}[t]
    \centering
    \includegraphics[width=\linewidth]{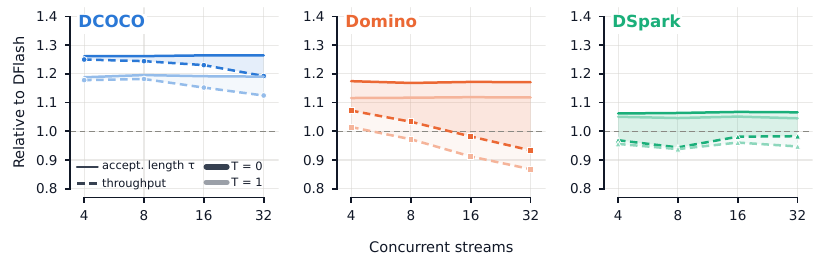}

    \caption{\textbf{At high concurrency, overhead outweighs acceptance
gains for Domino and DSpark, while \methodname{} remains faster
than DFlash.}
Accepted length $\tau$ (solid) and throughput (dashed), relative
to DFlash; Shading highlights the separation between the accepted-length and throughput ratios.
Qwen3-8B, thinking off, $T=0$ (dark) and $T=1$ (light),
averaged over all ten benchmark suites. Absolute values:
Tab.~\ref{tab:concurrency}.}
    \label{fig:concurrency}
\end{figure}

LLM serving systems batch concurrent requests to improve GPU utilization and throughput.
We therefore examine whether the benefits of draft correction persist as serving load
increases. Fig.~\ref{fig:q8b_conc} reports aggregate throughput from 4 to 32 concurrent
streams on Qwen3-8B with thinking off, averaged over all ten benchmark suites.
\methodname achieves the highest throughput at every concurrency, with margins over
the strongest alternative reaching $23\%$ under greedy decoding and $18\%$ under
sampling, and remaining $19\%$ and $12\%$, respectively, at 32 streams. The comparison with DFlash in Fig.~\ref{fig:concurrency} reveals how these gains evolve
under load. Although accepted length remains nearly unchanged, Domino and DSpark
progressively lose their throughput advantage and both become slower than the
uncorrected drafter at the highest concurrency as their correction overhead outweighs the benefit in this regime. In contrast,
\methodname remains faster than DFlash throughout, preserving the benefit of
correction even when the sequential alternatives incur a net slowdown.
Additional results for the distinct verifiers are reported in App.~\ref{app:concurrency}.

\subsection{Path selection strategy}
\label{sec:exp:path}

We compare three selection strategies over the same grid of transition scores produced by a trained corrector. Global-Sum maximizes the sum of transition scores across the block, while prefix-first greedy(Eq.~\ref{eq:greedy}) selects each successor using only the scores associated with the chosen predecessor. To test whether emphasizing earlier transitions improves global selection, we also evaluate a geometrically discounted variant,
\begin{equation}
\label{eq:geom}
\hat k_{1:B} \;=\; \arg\max_{k_1,\dots,k_B}\;
\Bigl[\, \ell_1(k_1) \;+\; \sum_{p=1}^{B-1} \lambda^{\,p-1}\,
S_p(k_p, k_{p+1}) \,\Bigr]\,,
\qquad \textsc{geom}(\lambda)\,,
\end{equation}
which keeps the global argmax but shifts weight toward earlier transitions; $\lambda = 1$ recovers the plain sum. We report $\lambda = 0.75$.
 
\begin{table}[t]
\centering
\fontsize{7.5}{8.5}\selectfont
\setlength{\tabcolsep}{2pt}

\begin{tabular*}{\linewidth}{@{\extracolsep{\fill}}llcccccc@{}}
\toprule
& & \multicolumn{3}{c}{Accepted length $\tau$}
  & \multicolumn{3}{c}{Throughput (tok/s)} \\
\cmidrule(lr){3-5}\cmidrule(lr){6-8}
Domain & Dataset & Greedy & Geometric & Global-Sum
 & Greedy & Geometric & Global-Sum \\
\midrule
\multirow{3}{*}{Math}
 & MATH-500      & 8.039 & \best{8.053} & 7.999 & \best{1627.2} & 1543.6 & 1565.9 \\
 & AIME-25       & 7.263 & \best{7.292} & 7.226 & \best{1437.7} & 1404.1 & 1392.3 \\
 & GSM8K         & \best{6.118} & 6.115 & 6.075 & \best{1176.9} & 1160.5 & 1169.4 \\
\midrule
\multirow{3}{*}{Code}
 & HumanEval     & 5.443 & \best{5.445} & 5.401 & \best{1077.0} & 1051.8 & 1059.0 \\
 & MBPP          & 5.215 & \best{5.231} & 5.181 & \best{979.3} & 971.1 & 969.7 \\
 & LiveCodeBench & \best{6.982} & 6.980 & 6.930 & \best{1326.4} & 1282.4 & 1275.5 \\
\midrule
\multirow{3}{*}{General}
 & MT-Bench      & \best{2.930} & 2.913 & 2.902 & \best{498.1} & 487.9 & 488.3 \\
 & Arena-Hard-v2 & 2.736 & \best{2.738} & 2.731 & \best{484.8} & 470.9 & 477.3 \\
 & Alpaca        & \best{2.560} & 2.551 & 2.543 & \best{497.1} & 476.9 & 484.2 \\
\midrule
\multicolumn{2}{@{}l}{Average}
 & 5.254 & \best{5.258} & 5.221 & \best{1011.6} & 983.2 & 986.8 \\
\bottomrule
\end{tabular*}

\caption{\textbf{Path selection over identical trained scores}
(Qwen3-8B from-pretrained corrector, $T{=}0$, no thinking,
2048 max tokens). Bold: best rule per row and metric.
Between greedy and geometric the $\tau$ differences carry
both signs and are an order of magnitude smaller than either
rule's gap to Global-Sum; the throughput ordering is uniform.}
\label{tab:walks}
\vspace{-4mm}
\end{table}
 
Tab.~\ref{tab:walks} shows that Global-Sum yields lower accepted length and throughput than Greedy on all nine evaluated suites. Geometric improves accepted length over Global-Sum on every suite and achieves acceptance comparable to Greedy: their per-suite differences have both signs, with average accepted lengths of 5.258 and 5.254, respectively. Greedy nevertheless achieves higher throughput on every suite, averaging a 2.8\% improvement over Geometric. These results support greedy selection as an effective acceptance–latency trade-off for the learned scores evaluated here, with an $(O(BK))$ walk and no discount hyperparameter.

\section{Conclusion}
\label{sec:conclusion}
Parallel block drafters leave two things on the table: their top-$K$
lists provide a significant coverage boost over the top-1 choice, and the identity of each
drafted token is information no other slot ever gets to use.
\methodname{} is a lightweight corrector built on exactly these two
facts. Reading only what the drafter already produced, it scores every
transition between adjacent candidates in parallel with a cheap neural
scorer, then greedily takes the best transition at each step to form
the proposed continuation, mirroring the way verification consumes
a draft, front to back, stopping at the first miss. Across several
backbone models and ten benchmarks, the corrected system leads in both
accepted length and decoding throughput at almost every point of
measurement, and its advantage persists or widens at higher serving concurrencies, where competing correctors fall below the uncorrected drafter.
Correction, made cheap enough, is pure profit for speculative
decoding.

\bibliography{references}
\bibliographystyle{iclr2027_conference}

\appendix

\section{Training details}
\label{app:training}

\paragraph{Training loss.}
For a single block the correction term is a weight-normalised cross-entropy over the
$K^2$ jointly softmax-normalised transition scores,
\begin{equation}
\label{eq:loss}
\mathcal{L}_{\mathrm{corr}}
\;=\;
-\frac{1}{Z}\sum_{p=1}^{B-1} e^{-(p-1)/\rho}\; m_p\,
\log \frac{\exp S_p(a^{\star}_p, b^{\star}_p)}
          {\sum_{a,b=1}^{K} \exp S_p(a,b)},
\qquad
Z = \sum_{p=1}^{B-1} e^{-(p-1)/\rho}\, m_p,
\quad \rho = 4,
\end{equation}
where the label $(a^{\star}_p, b^{\star}_p)$ is the pair of
verifier-argmax tokens at the two slots, computed over the
teacher-forced context. The mask $m_p \in \{0,1\}$ is $1$ only when
both verifier tokens lie inside their slots' top-$K$ candidate sets
and both slots pass the alignment mask; transitions the drafter cannot reach are dropped from supervision rather than taught. In a batch, the sums in
\eqref{eq:loss} and $Z$ extend jointly over all block, transition
pairs, a single weighted mean, not a mean of per-block means,
and $Z$ is clamped below by $1$, which bites only when a block has no
supervised transition. The complete loss is given by the adding the correction loss, with unit weight, to DFlash's own loss.

\paragraph{Training data.}
Following DFlash, we use approximately 800K prompts from the stem, math,
and code splits of the Nemotron Post-Training Dataset
v2~\citep{nvidia2025nvidianemotronnano2} and
CodeAlpaca-20k~\citep{codealpaca}. Each verifier regenerates its own
continuations using greedy decoding ($T=0$), with thinking disabled and
at most 4096 generated tokens. For each verifier, all methods train on
the same resulting corpus.

\paragraph{Training protocol.}
For each verifier, \methodname{} is initialized from a DFlash drafter
trained for six epochs, then trained for six further epochs alongside
the corrector. Baseline results use six training epochs under each method's published
objective; comparisons with extended training are reported in
Appendix~\ref{app:budget}.

\paragraph{Optimization and checkpoint selection.}
All methods use AdamW with learning rate $6\times10^{-4}$, no weight decay,
and a cosine learning-rate schedule with 4\% warmup. Training uses token-budgeted
sequence packing with a budget of 4096 tokens per GPU per micro-step across
eight GPUs and 23 gradient-accumulation steps. This corresponds to a nominal
budget of $4096\times8\times23=753{,}664$ tokens per optimizer step; the number
of sequences varies with their lengths. Validation metrics
are diagnostic and are not used for checkpoint selection. Each method is
trained once with seed 42, so variability across training seeds is not measured.
The drafter loss used by \methodname is specified above; baselines use their
published losses.

\paragraph{Relation of \eqref{eq:loss} to the decode rule.}
Because the label fixes the true predecessor, the joint cross-entropy
orders the true successor above every alternative in the predecessor's
row, precisely the conditional decision that
\eqref{eq:greedy} executes at decode time, while additionally
suppressing every competing row, so the training objective is strictly
stronger than the decision rule it must support. The exponential training weight emphasizes earlier transitions, motivated by the requirement that preceding tokens must be accepted before a later transition can contribute. It serves as a fixed proxy for prefix survival rather than an estimated acceptance probability. This training weight is distinct from the geometric discount evaluated in Section 5.3: the former controls the relative contribution of positions to the training loss, while the latter changes path selection at inference.
 
\paragraph{Label extraction.}
Training does not execute an online speculative loop. The context is a fixed,
teacher-forced corpus; the verifier's hidden states over it are
precomputed once, its output head is applied under no gradient, and
the argmax at each slot, when present in that slot's top-$K$,
supplies the label. Supervision is therefore conditioned exclusively
on verifier-correct prefixes. This is not a train/serve mismatch but a
consequence of prefix acceptance: a transition's decision is only ever
consumed when every earlier slot was accepted, i.e.\ matches the
verifier, so verifier-correct prefixes are precisely the states in
which the corrector's output matters.

\paragraph{Corrector architecture.}
The corrector uses candidate-representation width $d_e=128$, pair-state
width $d_h=128$, and $K=16$ candidates per drafted slot. These widths correspond
to $r=128$ in the method description. The two-layer scoring MLP has input
width $d_h+2d_e+2=386$, hidden width 256 with GELU activation, and a scalar
output. The pair projection has shape $128\times(2d)$ and the candidate--context
fusion projection has shape $128\times256$. On Qwen3-8B, the five-layer drafter
has hidden width $d=4096$ and the corrector has 2,229,505 parameters. The same
corrector architecture has 1,443,073 parameters on Qwen3-4B, reflecting the
smaller input hidden width. The frozen token-embedding table is shared with
the underlying model and is not an additional corrector parameter allocation.
Inference uses BF16 weights.

\paragraph{Draft-length convention.}
The nominal block-size setting is 16 for all methods $B=15$ refers to \methodname's drafted positions, excluding the anchor supplied
as context. \methodname therefore scores $(B-1)K^2=14\times16^2=3584$ adjacent
candidate transitions and commits at most 16 tokens per round, including the verifier's additional token. The comparisons use these native configurations rather than a strictly matched number of drafted tokens.

\section{Training budget and convergence}
\label{app:budget}
 
Our arm trains for twelve epochs in total, six drafter-only,
matching the published DFlash recipe, followed by six joint epochs
with the corrector, while every baseline trains for its standard
six. To rule out budget rather than the corrector as the source of
the gains, we continue each baseline (DFlash, Domino, DSpark) to
twelve epochs under its own objective and recipe, unchanged.
Table~\ref{tab:budget} reports the result: all three are converged at
six epochs, doubling their training moves accepted length by
no more than 1.9\%, whereas the joint phase is where our arm's gain
accrues. The comparison in the main tables is therefore between
converged systems, not between budgets.
 
\begin{table}[h]
\centering
\small
\begin{tabular}{lcccc}
\toprule
Arm & $\tau$ @ 6 epochs & $\tau$ @ 12 epochs & $\Delta\tau$ &
$\Delta$TPS \\
\midrule
DFlash & 3.913 & 3.948 & +0.035 (+0.9\%)& +11.0 (+1.4\%) \\
Domino & 4.613 & 4.701 & +0.088 (+1.9\%) & +10.9 (+1.3\%) \\
DSpark & 4.254 & 4.321 & +0.067 (+1.6\%) & +3.1 (+0.4\%) \\
\midrule
\methodname (6 + 6 joint) & 3.913 {\scriptsize(drafter-only @ 6)} & 4.985 &
+1.072 (+27.4\%) & +176.6 (+22.6\%) \\
\bottomrule
\end{tabular}
\caption{\textbf{Extending baseline training to match our total
budget}, Q8B no-thinking, max generated tokens 2048, T=0 averaged across the 10 benchmarks. Baselines show no more than 2\% improvement from a second six epochs; the joint phase accounts for our arm's improvement.
}
\label{tab:budget}
\end{table}

\paragraph{Component ablations.}
The corrector's correction term $\Delta_\theta$ reads two learned
input pathways: the candidate representations $z^a_p, z^b_{p+1}$ and
the candidate-independent pair state $s_p$. We retrain the corrector
with each pathway removed, under the recipe of
App.~\ref{app:training} unchanged: the $-$pair state variant drops the pair state
(transitions scored from candidate representations and logits alone)
and the $-$cand state variant drops the candidate representations (a candidate-blind
scorer that recalibrates the draft logits from the pair state alone).
The Frozen backbone ablation keeps the corrector intact but
freezes the drafter at its six-epoch DFlash checkpoint, so the
corrector trains as a pure drop-in and the backbone never co-adapts.
Tables~\ref{tab:ablations} and~\ref{tab:ablations-tau} report throughput
and accepted length, respectively.
Removing either input pathway hurts both metrics across all ten suites
and both temperatures. The pair state carries the larger
share: without it, accepted length drops by $7.3\%$/$5.6\%$
($T{=}0$/$T{=}1$ averages) and throughput by $6.3\%$/$4.6\%$; without
candidate representations, accepted length drops by $4.8\%$/$3.7\%$
and throughput by $4.6\%$/$3.3\%$. Neither pathway is redundant, and
their contributions are consistent across domains, largest on math
and code. Even with the drafter frozen, training the corrector increases
average throughput from $783.1$ to $817.7$~tok/s at $T{=}0$ ($+4.4\%$).
Freezing the drafter nevertheless reduces average throughput by
$15.9\%$/$12.9\%$ at $T{=}0$/$T{=}1$, respectively, relative to joint
training. By comparison, extending drafter training alone improves
throughput by at most $2\%$ (Tab.~\ref{tab:budget}), suggesting that
the gains from joint training cannot be explained by additional
drafter training alone.

\begin{table}[t]
\centering
\small
\begin{minipage}{\linewidth}
\centering
\textbf{(a)} Greedy decoding, $T{=}0$ (mean of 3 runs)\\[3pt]
\resizebox{\linewidth}{!}{%
\begin{tabular}{llcccc}
\toprule
Domain & Dataset & \methodname & $-$pair state & $-$cand state & Frozen backbone \\
\midrule
\multirow{3}{*}{Math} & MATH-500 & \best{1599.3} & 1497.7 ($-$6.4\%) & 1521.8 ($-$4.8\%) & 1333.6 ($-$16.6\%) \\
 & AIME-25 & \best{1475.6} & 1382.0 ($-$6.3\%) & 1387.9 ($-$5.9\%) & 1236.1 ($-$16.2\%) \\
 & GSM8K & \best{1130.3} & 1042.3 ($-$7.8\%) & 1073.2 ($-$5.0\%) & 920.2 ($-$18.6\%) \\
\midrule
\multirow{3}{*}{Code} & HumanEval & \best{1113.9} & 1035.6 ($-$7.0\%) & 1055.9 ($-$5.2\%) & 929.5 ($-$16.6\%) \\
 & MBPP & \best{1121.6} & 1048.7 ($-$6.5\%) & 1067.8 ($-$4.8\%) & 929.5 ($-$17.1\%) \\
 & LiveCodeBench & \best{1202.9} & 1121.2 ($-$6.8\%) & 1139.7 ($-$5.3\%) & 1019.2 ($-$15.3\%) \\
\midrule
\multirow{4}{*}{General} & MT-Bench & \best{575.2} & 548.4 ($-$4.7\%) & 565.2 ($-$1.8\%) & 499.6 ($-$13.2\%) \\
 & Arena-Hard-v2 & \best{516.2} & 493.4 ($-$4.4\%) & 504.7 ($-$2.2\%) & 450.1 ($-$12.8\%) \\
 & Alpaca & \best{485.2} & 461.5 ($-$4.9\%) & 472.2 ($-$2.7\%) & 429.3 ($-$11.5\%) \\
 & SpeedBench & \best{498.0} & 475.8 ($-$4.5\%) & 481.1 ($-$3.4\%) & 430.3 ($-$13.6\%) \\
\midrule
\multicolumn{2}{l}{Average} & \best{971.8} & 910.7 ($-$6.3\%) & 927.1 ($-$4.6\%) & 817.7 ($-$15.9\%) \\
\bottomrule
\end{tabular}%
}
\end{minipage}\\[10pt]
\begin{minipage}{\linewidth}
\centering
\textbf{(b)} Sampling, $T{=}1$ (mean{\scriptsize$\pm$}std over seeds 42/43/44)\\[3pt]
\resizebox{\linewidth}{!}{%
\begin{tabular}{llcccc}
\toprule
Domain & Dataset & \methodname & $-$pair state & $-$cand state & Frozen backbone \\
\midrule
\multirow{3}{*}{Math} & MATH-500 & \best{\pmstd{1133.8}{5.0}} & \pmstd{1094.1}{3.5} ($-$3.5\%) & \pmstd{1101.1}{6.5} ($-$2.9\%) & \pmstd{988.6}{10.2} ($-$12.8\%) \\
 & AIME-25 & \best{\pmstd{884.8}{13.7}} & \pmstd{848.2}{22.9} ($-$4.1\%) & \pmstd{859.7}{6.4} ($-$2.8\%) & \pmstd{785.3}{13.2} ($-$11.2\%) \\
 & GSM8K & \best{\pmstd{1007.8}{18.7}} & \pmstd{940.7}{6.0} ($-$6.7\%) & \pmstd{971.8}{11.5} ($-$3.6\%) & \pmstd{837.5}{10.5} ($-$16.9\%) \\
\midrule
\multirow{3}{*}{Code} & HumanEval & \best{\pmstd{945.3}{4.4}} & \pmstd{882.5}{1.4} ($-$6.6\%) & \pmstd{901.9}{8.0} ($-$4.6\%) & \pmstd{806.4}{9.6} ($-$14.7\%) \\
 & MBPP & \best{\pmstd{894.9}{6.4}} & \pmstd{852.4}{2.4} ($-$4.7\%) & \pmstd{855.2}{3.5} ($-$4.4\%) & \pmstd{771.8}{2.7} ($-$13.8\%) \\
 & LiveCodeBench & \best{\pmstd{766.5}{9.5}} & \pmstd{736.2}{4.2} ($-$4.0\%) & \pmstd{738.2}{5.2} ($-$3.7\%) & \pmstd{678.2}{6.0} ($-$11.5\%) \\
\midrule
\multirow{4}{*}{General} & MT-Bench & \best{\pmstd{507.9}{1.5}} & \pmstd{492.1}{3.3} ($-$3.1\%) & \pmstd{496.9}{5.2} ($-$2.2\%) & \pmstd{456.3}{4.8} ($-$10.2\%) \\
 & Arena-Hard-v2 & \best{\pmstd{439.7}{3.0}} & \pmstd{422.5}{1.4} ($-$3.9\%) & \pmstd{430.9}{1.5} ($-$2.0\%) & \pmstd{393.4}{1.5} ($-$10.5\%) \\
 & Alpaca & \best{\pmstd{444.2}{1.5}} & \pmstd{425.6}{2.0} ($-$4.2\%) & \pmstd{432.6}{3.0} ($-$2.6\%) & \pmstd{402.0}{2.4} ($-$9.5\%) \\
 & SpeedBench & \best{\pmstd{460.0}{16.9}} & \pmstd{442.5}{3.4} ($-$3.8\%) & \pmstd{447.0}{5.3} ($-$2.8\%) & \pmstd{398.5}{2.7} ($-$13.4\%) \\
\midrule
\multicolumn{2}{l}{Average} & \best{748.5} & 713.7 ($-$4.6\%) & 723.5 ($-$3.3\%) & 651.8 ($-$12.9\%) \\
\bottomrule
\end{tabular}%
}
\end{minipage}
\caption{\textbf{Ablations: throughput (output tokens/s, per-stream),
Qwen3-8B, thinking off}, concurrency~1, max 2048 generated tokens,
all ten suites. Each ablation cell reports its value with the change
relative to \methodname{} in parentheses.
$-$pair state removes the corrector's dedicated previous-state channel;
$-$cand state zeroes the candidate representations entering the transition scorer;
Frozen backbone trains only the corrector on the fixed six-epoch DFlash backbone.
Bold: best measured system per row.}
\label{tab:ablations}
\end{table}

\begin{table}[t]
\centering
\small
\begin{minipage}{\linewidth}
\centering
\textbf{(a)} Greedy decoding, $T{=}0$ (mean of 3 runs)\\[3pt]
\resizebox{\linewidth}{!}{%
\begin{tabular}{llcccc}
\toprule
Domain & Dataset & \methodname & $-$pair state & $-$cand state & Frozen backbone \\
\midrule
\multirow{3}{*}{Math} & MATH-500 & \best{8.256} & 7.588 ($-$8.1\%) & 7.802 ($-$5.5\%) & 6.801 ($-$17.6\%) \\
 & AIME-25 & \best{7.506} & 6.935 ($-$7.6\%) & 7.043 ($-$6.2\%) & 6.250 ($-$16.7\%) \\
 & GSM8K & \best{6.010} & 5.482 ($-$8.8\%) & 5.697 ($-$5.2\%) & 4.845 ($-$19.4\%) \\
\midrule
\multirow{3}{*}{Code} & HumanEval & \best{5.746} & 5.340 ($-$7.1\%) & 5.439 ($-$5.3\%) & 4.764 ($-$17.1\%) \\
 & MBPP & \best{5.858} & 5.379 ($-$8.2\%) & 5.575 ($-$4.8\%) & 4.800 ($-$18.1\%) \\
 & LiveCodeBench & \best{6.153} & 5.712 ($-$7.2\%) & 5.790 ($-$5.9\%) & 5.196 ($-$15.5\%) \\
\midrule
\multirow{4}{*}{General} & MT-Bench & \best{2.930} & 2.760 ($-$5.8\%) & 2.860 ($-$2.4\%) & 2.539 ($-$13.4\%) \\
 & Arena-Hard-v2 & \best{2.605} & 2.479 ($-$4.9\%) & 2.550 ($-$2.1\%) & 2.257 ($-$13.3\%) \\
 & Alpaca & \best{2.474} & 2.344 ($-$5.3\%) & 2.407 ($-$2.7\%) & 2.173 ($-$12.2\%) \\
 & SpeedBench & \best{2.561} & 2.434 ($-$5.0\%) & 2.499 ($-$2.4\%) & 2.207 ($-$13.8\%) \\
\midrule
\multicolumn{2}{l}{Average} & \best{5.010} & 4.645 ($-$7.3\%) & 4.767 ($-$4.8\%) & 4.183 ($-$16.5\%) \\
\bottomrule
\end{tabular}%
}
\end{minipage}\\[10pt]
\begin{minipage}{\linewidth}
\centering
\textbf{(b)} Sampling, $T{=}1$ (mean{\scriptsize$\pm$}std over seeds 42/43/44)\\[3pt]
\resizebox{\linewidth}{!}{%
\begin{tabular}{llcccc}
\toprule
Domain & Dataset & \methodname & $-$pair state & $-$cand state & Frozen backbone \\
\midrule
\multirow{3}{*}{Math} & MATH-500 & \best{\pmstd{5.801}{0.025}} & \pmstd{5.495}{0.019} ($-$5.3\%) & \pmstd{5.589}{0.037} ($-$3.7\%) & \pmstd{5.004}{0.052} ($-$13.7\%) \\
 & AIME-25 & \best{\pmstd{4.461}{0.070}} & \pmstd{4.235}{0.114} ($-$5.1\%) & \pmstd{4.318}{0.039} ($-$3.2\%) & \pmstd{3.951}{0.068} ($-$11.4\%) \\
 & GSM8K & \best{\pmstd{5.324}{0.104}} & \pmstd{4.912}{0.048} ($-$7.7\%) & \pmstd{5.089}{0.061} ($-$4.4\%) & \pmstd{4.392}{0.057} ($-$17.5\%) \\
\midrule
\multirow{3}{*}{Code} & HumanEval & \best{\pmstd{4.839}{0.022}} & \pmstd{4.527}{0.006} ($-$6.5\%) & \pmstd{4.618}{0.031} ($-$4.6\%) & \pmstd{4.109}{0.050} ($-$15.1\%) \\
 & MBPP & \best{\pmstd{4.636}{0.033}} & \pmstd{4.339}{0.014} ($-$6.4\%) & \pmstd{4.422}{0.023} ($-$4.6\%) & \pmstd{3.962}{0.016} ($-$14.5\%) \\
 & LiveCodeBench & \best{\pmstd{3.891}{0.050}} & \pmstd{3.714}{0.010} ($-$4.5\%) & \pmstd{3.719}{0.026} ($-$4.4\%) & \pmstd{3.436}{0.032} ($-$11.7\%) \\
\midrule
\multirow{4}{*}{General} & MT-Bench & \best{\pmstd{2.581}{0.007}} & \pmstd{2.471}{0.017} ($-$4.3\%) & \pmstd{2.507}{0.026} ($-$2.9\%) & \pmstd{2.315}{0.023} ($-$10.3\%) \\
 & Arena-Hard-v2 & \best{\pmstd{2.218}{0.015}} & \pmstd{2.119}{0.007} ($-$4.4\%) & \pmstd{2.175}{0.009} ($-$2.0\%) & \pmstd{1.972}{0.008} ($-$11.1\%) \\
 & Alpaca & \best{\pmstd{2.265}{0.009}} & \pmstd{2.161}{0.010} ($-$4.6\%) & \pmstd{2.205}{0.014} ($-$2.6\%) & \pmstd{2.034}{0.012} ($-$10.2\%) \\
 & SpeedBench & \best{\pmstd{2.373}{0.101}} & \pmstd{2.252}{0.024} ($-$5.1\%) & \pmstd{2.313}{0.026} ($-$2.5\%) & \pmstd{2.050}{0.011} ($-$13.6\%) \\
\midrule
\multicolumn{2}{l}{Average} & \best{3.839} & 3.623 ($-$5.6\%) & 3.695 ($-$3.7\%) & 3.322 ($-$13.5\%) \\
\bottomrule
\end{tabular}%
}
\end{minipage}
\caption{\textbf{Ablations: accepted length $\tau$,
Qwen3-8B, thinking off}, concurrency~1, max 2048 generated tokens,
all ten suites. Each ablation cell reports its value with the change
relative to \methodname{} in parentheses.
$-$pair state removes the corrector's dedicated previous-state channel;
$-$cand state zeroes the candidate representations entering the transition scorer;
Frozen backbone trains only the corrector on the fixed six-epoch DFlash backbone.
Bold: best measured system per row.}
\label{tab:ablations-tau}
\end{table}

\clearpage
\section{Evaluation details}
\label{app:eval}

\paragraph{Benchmarks and generation settings.}
We evaluate ten benchmarks across three domains: mathematics
(GSM8K~\citep{cobbe2021gsm8k}, MATH-500~\citep{lightman2024let}, AIME-25~\citep{aime25}), code (HumanEval~\citep{chen2021codex}, MBPP~\citep{austin2021program}, LiveCodeBench~\citep{jain2025livecodebench}), and
general tasks and chat (SpeedBench~\citep{abramovich2026speedbench}, Alpaca~\citep{alpaca}, MT-Bench~\citep{zheng2023judging}, Arena-Hard-v2~\citep{li2025arenahard}).
We evaluate greedy decoding at $T=0$ and stochastic decoding at $T=1$.
The maximum number of generated tokens is 2048 without thinking and
4096 with extended thinking. Both thinking settings are evaluated for
Qwen3-4B and Qwen3-8B; Qwen3-Coder-30B-A3B and Llama-3.1-8B are
evaluated without a separate thinking mode.

\paragraph{Serving configuration.}
All systems run on a single NVIDIA B200 GPU with tensor parallelism 1,
using vLLM 0.27.0 and PyTorch 2.13.0+cu130. Model precision is BF16
(\texttt{--dtype auto}); KV-cache dtype uses the automatic setting
(\texttt{--kv-cache-dtype auto}). Prefix caching is disabled
(\texttt{--no-enable-prefix-caching}). CUDA Graphs are enabled, eager
execution is not forced, and no optimization-level override is supplied.
Evaluation begins after server initialization and CUDA Graph capture complete.

\paragraph{Timing and aggregation.}
For request $i$, let $n_i$ denote the number of generated output tokens and
$L_i$ the request latency in seconds, including prefill. Its reported rate is
$n_i/L_i$, rather than the inverse inter-token latency. At concurrency 1,
we report the median request-level rate within each benchmark. The reported
ten-suite averages give each benchmark equal weight. No initial evaluation
requests are discarded as warmup. For $T=1$, the reported mean and standard
deviation are computed across generation seeds 42, 43, and 44, after computing
the benchmark statistic for each seed. These quantify evaluation variability
for fixed checkpoints, not variability across independent training runs.

\paragraph{Evaluation subsets.}
We use fixed benchmark subsets selected with subsampling seed 1234, shared
across methods. GSM8K, MATH-500, HumanEval, MBPP, LiveCodeBench, Alpaca, and
Arena-Hard-v2 each contain 128 selected rows. AIME-25 uses all 30 rows and
MT-Bench all 80 rows. SPEED-Bench contains 80 rows from each of 11 subsets,
for 880 rows in total, and is treated as one of the ten benchmark suites.
The evaluation manifest records the exact selected row indices.
For LiveCodeBench, we sample from the 1,055 available rows in the
\texttt{code\_generation\_lite} v5 and v6 release files,
\texttt{test5.jsonl} and \texttt{test6.jsonl}
(\texttt{version\_tag=v5\_v6}).

\section{Full per-dataset results}
\label{app:full-results}

Per-dataset throughput and accepted length for each configuration not
shown in the main text, in the format of Tables~\ref{tab:main-q8b}
and~\ref{tab:tau-q8b}.

\begin{table}[t]
\centering
\small
\setlength{\tabcolsep}{3.5pt}
\resizebox{\linewidth}{!}{%
\begin{tabular}{llcccccccc}
\toprule
& & \multicolumn{2}{c}{DFlash} & \multicolumn{2}{c}{Domino} &
\multicolumn{2}{c}{DSpark} & \multicolumn{2}{c}{\methodname} \\
\cmidrule(lr){3-4}\cmidrule(lr){5-6}\cmidrule(lr){7-8}\cmidrule(lr){9-10}
Domain & Dataset & $T{=}0$ & $T{=}1$ & $T{=}0$ & $T{=}1$
 & $T{=}0$ & $T{=}1$ & $T{=}0$ & $T{=}1$ \\
\midrule
\multirow{3}{*}{Math}
 & MATH-500       & 6.173 & \pmstd{4.751}{0.024} & 7.500 & \pmstd{5.518}{0.022} & 6.815 & \pmstd{5.245}{0.053} & \best{8.039} & \best{\pmstd{5.755}{0.073}} \\
 & AIME-25        & 5.554 & \pmstd{3.751}{0.064} & 6.803 & \pmstd{4.246}{0.076} & 6.289 & \pmstd{3.975}{0.059} & \best{7.263} & \best{\pmstd{4.486}{0.037}} \\
 & GSM8K          & 4.632 & \pmstd{4.217}{0.024} & 5.563 & \pmstd{4.960}{0.122} & 4.992 & \pmstd{4.526}{0.019} & \best{6.118} & \best{\pmstd{5.440}{0.141}} \\
\midrule
\multirow{3}{*}{Code}
 & HumanEval      & 4.245 & \pmstd{3.771}{0.009} & 4.971 & \pmstd{4.319}{0.075} & 4.576 & \pmstd{4.046}{0.033} & \best{5.443} & \best{\pmstd{4.646}{0.043}} \\
 & MBPP           & 3.992 & \pmstd{3.507}{0.014} & 4.857 & \pmstd{4.121}{0.117} & 4.345 & \pmstd{3.767}{0.009} & \best{5.215} & \best{\pmstd{4.428}{0.008}} \\
 & LiveCodeBench  & 5.530 & \pmstd{4.398}{0.140} & 6.503 & \pmstd{4.908}{0.083} & 6.342 & \pmstd{4.920}{0.089} & \best{6.982} & \best{\pmstd{5.431}{0.160}} \\
\midrule
\multirow{4}{*}{General}
 & MT-Bench       & 2.444 & \pmstd{2.249}{0.010} & 2.736 & \pmstd{2.475}{0.044} & 2.456 & \pmstd{2.318}{0.016} & \best{2.930} & \best{\pmstd{2.581}{0.007}} \\
 & Arena-Hard-v2  & 2.288 & \pmstd{1.967}{0.029} & 2.473 & \pmstd{2.101}{0.036} & 2.288 & \pmstd{2.019}{0.009} & \best{2.736} & \best{\pmstd{2.286}{0.020}} \\
 & Alpaca         & 2.147 & \pmstd{1.983}{0.025} & 2.393 & \pmstd{2.169}{0.026} & 2.242 & \pmstd{2.051}{0.009} & \best{2.560} & \best{\pmstd{2.313}{0.024}} \\
 & SpeedBench      & 2.124 & \pmstd{1.994}{0.032} & 2.327 & \pmstd{2.156}{0.036} & 2.195 & \pmstd{2.070}{0.016} & \best{2.561} & \best{\pmstd{2.373}{0.101}} \\
\midrule
\multicolumn{2}{l}{Average} & 3.913 & 3.259 & 4.613 & 3.697 & 4.254 & 3.494 & \best{4.985} & \best{3.974} \\
\bottomrule
\end{tabular}%
}
\caption{\textbf{Mechanism: mean accepted length $\tau$, Qwen3-8B,
thinking off}, companion to Table~\ref{tab:main-q8b}; no-spec commits
one token per round by construction and is omitted. $\tau$ explains
the TPS movement; the corrector's own overhead is visible only in
throughput.}
\label{tab:tau-q8b}
\end{table}

\begin{table}[h]
\centering
\small
\begin{minipage}{\linewidth}
\centering
\textbf{(a)} Greedy decoding, $T{=}0$ (single deterministic run)\\[3pt]
\resizebox{\linewidth}{!}{%
\begin{tabular}{llccccc}
\toprule
Domain & Dataset & No-spec & DFlash & Domino & DSpark & \methodname \\
\midrule
\multirow{3}{*}{Math}
 & MATH-500       & 262.0 & 608.8 (2.32$\times$) & 617.1 (2.36$\times$) & 567.4 (2.17$\times$) & \best{724.9 (2.77$\times$)} \\
 & AIME-25        & 260.3 & 586.9 (2.26$\times$) & 612.5 (2.35$\times$) & 574.9 (2.21$\times$) & \best{703.1 (2.70$\times$)} \\
 & GSM8K          & 264.1 & 555.6 (2.10$\times$) & 559.7 (2.12$\times$) & 518.6 (1.96$\times$) & \best{646.5 (2.45$\times$)} \\
\midrule
\multirow{3}{*}{Code}
 & HumanEval      & 260.6 & 562.9 (2.16$\times$) & 572.8 (2.20$\times$) & 524.7 (2.01$\times$) & \best{654.4 (2.51$\times$)} \\
 & MBPP           & 262.8 & 535.6 (2.04$\times$) & 548.2 (2.09$\times$) & 498.2 (1.90$\times$) & \best{618.1 (2.35$\times$)} \\
 & LiveCodeBench  & 259.2 & 541.9 (2.09$\times$) & 554.5 (2.14$\times$) & 506.6 (1.95$\times$) & \best{624.2 (2.41$\times$)} \\
\midrule
\multirow{4}{*}{General}
 & MT-Bench       & 264.4 & 477.4 (1.81$\times$) & 481.9 (1.82$\times$) & 465.3 (1.76$\times$) & \best{545.6 (2.06$\times$)} \\
 & Arena-Hard-v2  & 260.4 & 439.1 (1.69$\times$) & 421.9 (1.62$\times$) & 404.8 (1.56$\times$) & \best{508.3 (1.95$\times$)} \\
 & Alpaca         & 265.8 & 388.1 (1.46$\times$) & 389.0 (1.46$\times$) & 364.2 (1.37$\times$) & \best{440.8 (1.66$\times$)} \\
 & SpeedBench      & 264.3 & 403.3 (1.53$\times$) & 403.3 (1.53$\times$) & 380.9 (1.44$\times$) & \best{458.7 (1.74$\times$)} \\
\midrule
\multicolumn{2}{l}{Average} & 262.4 & 510.0 (1.95$\times$) & 516.1 (1.97$\times$) & 480.6 (1.83$\times$) & \best{592.5 (2.26$\times$)} \\
\bottomrule
\end{tabular}%
}
\end{minipage}\\[10pt]
\begin{minipage}{\linewidth}
\centering
\textbf{(b)} Sampling, $T{=}1$ (mean{\scriptsize$\pm$}std over seeds 42/43/44)\\[3pt]
\resizebox{\linewidth}{!}{%
\begin{tabular}{llccccc}
\toprule
Domain & Dataset & No-spec & DFlash & Domino & DSpark & \methodname \\
\midrule
\multirow{3}{*}{Math}
 & MATH-500       & \pmstd{261.8}{0.0} & \pmstd{567.0}{3.3} (2.17$\times$) & \pmstd{566.2}{5.3} (2.16$\times$) & \pmstd{538.8}{3.0} (2.06$\times$) & \best{\pmstd{664.1}{7.6} (2.54$\times$)} \\
 & AIME-25        & \pmstd{259.7}{0.0} & \pmstd{532.1}{4.7} (2.05$\times$) & \pmstd{570.5}{13.0} (2.20$\times$) & \pmstd{514.4}{17.1} (1.98$\times$) & \best{\pmstd{629.3}{15.4} (2.42$\times$)} \\
 & GSM8K          & \pmstd{264.0}{0.1} & \pmstd{528.1}{3.6} (2.00$\times$) & \pmstd{522.7}{1.2} (1.98$\times$) & \pmstd{495.6}{5.0} (1.88$\times$) & \best{\pmstd{599.6}{6.2} (2.27$\times$)} \\
\midrule
\multirow{3}{*}{Code}
 & HumanEval      & \pmstd{259.9}{0.1} & \pmstd{502.0}{3.1} (1.93$\times$) & \pmstd{508.6}{5.4} (1.96$\times$) & \pmstd{479.4}{4.9} (1.85$\times$) & \best{\pmstd{572.4}{8.6} (2.20$\times$)} \\
 & MBPP           & \pmstd{262.3}{0.2} & \pmstd{479.1}{1.0} (1.83$\times$) & \pmstd{484.1}{7.1} (1.85$\times$) & \pmstd{446.2}{0.9} (1.70$\times$) & \best{\pmstd{541.7}{6.6} (2.07$\times$)} \\
 & LiveCodeBench  & \pmstd{258.6}{0.0} & \pmstd{474.2}{0.2} (1.83$\times$) & \pmstd{479.5}{3.8} (1.85$\times$) & \pmstd{446.6}{2.0} (1.73$\times$) & \best{\pmstd{535.4}{2.0} (2.07$\times$)} \\
\midrule
\multirow{4}{*}{General}
 & MT-Bench       & \pmstd{264.2}{0.1} & \pmstd{427.2}{9.8} (1.62$\times$) & \pmstd{436.6}{4.7} (1.65$\times$) & \pmstd{404.5}{8.6} (1.53$\times$) & \best{\pmstd{475.4}{7.7} (1.80$\times$)} \\
 & Arena-Hard-v2  & \pmstd{259.8}{0.0} & \pmstd{391.4}{5.5} (1.51$\times$) & \pmstd{383.9}{5.9} (1.48$\times$) & \pmstd{366.7}{5.6} (1.41$\times$) & \best{\pmstd{441.4}{13.3} (1.70$\times$)} \\
 & Alpaca         & \pmstd{265.6}{0.1} & \pmstd{374.9}{1.4} (1.41$\times$) & \pmstd{371.9}{7.3} (1.40$\times$) & \pmstd{352.2}{0.9} (1.33$\times$) & \best{\pmstd{421.5}{3.1} (1.59$\times$)} \\
 & SpeedBench      & \pmstd{263.9}{0.1} & \pmstd{375.8}{5.7} (1.42$\times$) & \pmstd{371.0}{5.7} (1.41$\times$) & \pmstd{344.6}{10.0} (1.31$\times$) & \best{\pmstd{423.2}{11.3} (1.60$\times$)} \\
\midrule
\multicolumn{2}{l}{Average} & 262.0 & 465.2 (1.78$\times$) & 469.5 (1.79$\times$) & 438.9 (1.68$\times$) & \best{530.4 (2.03$\times$)} \\
\bottomrule
\end{tabular}%
}
\end{minipage}
\caption{\textbf{Throughput (output tokens/s, per-stream) and speedup
over no-spec decoding, Qwen3-8B, thinking on}, concurrency~1, max
4096 generated tokens. Same format as Table~\ref{tab:main-q8b}.}
\label{tab:main-q8b-think}
\end{table}

\begin{table}[h]
\centering
\small
\setlength{\tabcolsep}{3.5pt}
\resizebox{\linewidth}{!}{%
\begin{tabular}{llcccccccc}
\toprule
& & \multicolumn{2}{c}{DFlash} & \multicolumn{2}{c}{Domino} &
\multicolumn{2}{c}{DSpark} & \multicolumn{2}{c}{\methodname} \\
\cmidrule(lr){3-4}\cmidrule(lr){5-6}\cmidrule(lr){7-8}\cmidrule(lr){9-10}
Domain & Dataset & $T{=}0$ & $T{=}1$ & $T{=}0$ & $T{=}1$
 & $T{=}0$ & $T{=}1$ & $T{=}0$ & $T{=}1$ \\
\midrule
\multirow{3}{*}{Math}
 & MATH-500       & 2.947 & \pmstd{2.759}{0.017} & 3.298 & \pmstd{3.057}{0.018} & 3.102 & \pmstd{2.884}{0.016} & \best{3.590} & \best{\pmstd{3.304}{0.021}} \\
 & AIME-25        & 2.887 & \pmstd{2.625}{0.035} & 3.250 & \pmstd{2.981}{0.015} & 3.039 & \pmstd{2.788}{0.021} & \best{3.526} & \best{\pmstd{3.170}{0.013}} \\
 & GSM8K          & 2.717 & \pmstd{2.560}{0.004} & 3.025 & \pmstd{2.820}{0.005} & 2.788 & \pmstd{2.649}{0.012} & \best{3.287} & \best{\pmstd{3.040}{0.015}} \\
\midrule
\multirow{3}{*}{Code}
 & HumanEval      & 2.786 & \pmstd{2.443}{0.004} & 3.099 & \pmstd{2.702}{0.016} & 2.876 & \pmstd{2.561}{0.004} & \best{3.331} & \best{\pmstd{2.879}{0.018}} \\
 & MBPP           & 2.588 & \pmstd{2.326}{0.003} & 2.892 & \pmstd{2.555}{0.020} & 2.681 & \pmstd{2.420}{0.009} & \best{3.093} & \best{\pmstd{2.703}{0.011}} \\
 & LiveCodeBench  & 2.631 & \pmstd{2.325}{0.010} & 2.934 & \pmstd{2.562}{0.014} & 2.720 & \pmstd{2.414}{0.015} & \best{3.152} & \best{\pmstd{2.715}{0.008}} \\
\midrule
\multirow{4}{*}{General}
 & MT-Bench       & 2.307 & \pmstd{2.096}{0.013} & 2.541 & \pmstd{2.288}{0.007} & 2.432 & \pmstd{2.163}{0.009} & \best{2.738} & \best{\pmstd{2.435}{0.016}} \\
 & Arena-Hard-v2  & 2.160 & \pmstd{1.905}{0.010} & 2.333 & \pmstd{2.045}{0.002} & 2.197 & \pmstd{1.968}{0.013} & \best{2.576} & \best{\pmstd{2.217}{0.029}} \\
 & Alpaca         & 1.982 & \pmstd{1.848}{0.007} & 2.153 & \pmstd{2.012}{0.028} & 2.028 & \pmstd{1.917}{0.020} & \best{2.314} & \best{\pmstd{2.146}{0.029}} \\
 & SpeedBench      & 2.008 & \pmstd{1.849}{0.026} & 2.193 & \pmstd{1.997}{0.025} & 2.085 & \pmstd{1.878}{0.043} & \best{2.380} & \best{\pmstd{2.159}{0.049}} \\
\midrule
\multicolumn{2}{l}{Average} & 2.501 & 2.274 & 2.772 & 2.502 & 2.595 & 2.364 & \best{2.999} & \best{2.677} \\
\bottomrule
\end{tabular}%
}
\caption{\textbf{Mean accepted length $\tau$, Qwen3-8B, thinking on},
companion to Table~\ref{tab:main-q8b-think}; \methodname attains the best
$\tau$ in every cell in this configuration.}
\label{tab:tau-q8b-think}
\end{table}

\begin{table}[h]
\centering
\small
\begin{minipage}{\linewidth}
\centering
\textbf{(a)} Greedy decoding, $T{=}0$ (single deterministic run)\\[3pt]
\resizebox{\linewidth}{!}{%
\begin{tabular}{llccccc}
\toprule
Domain & Dataset & No-spec & DFlash & Domino & DSpark & \methodname \\
\midrule
\multirow{3}{*}{Math}
 & MATH-500       & 278.4 & 1356.9 (4.88$\times$) & 1477.5 (5.31$\times$) & 1367.6 (4.91$\times$) & \best{1636.0 (5.88$\times$)} \\
 & AIME-25        & 274.8 & 1465.6 (5.33$\times$) & 1601.0 (5.83$\times$) & 1491.2 (5.43$\times$) & \best{1811.7 (6.59$\times$)} \\
 & GSM8K          & 278.4 & 891.0 (3.20$\times$) & 915.5 (3.29$\times$) & 865.8 (3.11$\times$) & \best{1160.8 (4.17$\times$)} \\
\midrule
\multirow{3}{*}{Code}
 & HumanEval      & 276.4 & 1049.9 (3.80$\times$) & 1144.7 (4.14$\times$) & 1038.0 (3.76$\times$) & \best{1260.3 (4.56$\times$)} \\
 & MBPP           & 282.3 & 1036.3 (3.67$\times$) & \best{1128.5 (4.00$\times$)} & 1039.0 (3.68$\times$) & 989.7 (3.51$\times$) \\
 & LiveCodeBench  & 278.7 & 990.0 (3.55$\times$) & 1050.5 (3.77$\times$) & 972.1 (3.49$\times$) & \best{1448.8 (5.20$\times$)} \\
\midrule
\multirow{4}{*}{General}
 & MT-Bench       & 281.6 & 420.8 (1.49$\times$) & 436.4 (1.55$\times$) & 416.9 (1.48$\times$) & \best{572.5 (2.03$\times$)} \\
 & Arena-Hard-v2  & 279.0 & 409.0 (1.47$\times$) & 407.3 (1.46$\times$) & 387.6 (1.39$\times$) & \best{570.5 (2.05$\times$)} \\
 & Alpaca         & 279.2 & 409.8 (1.47$\times$) & 426.7 (1.53$\times$) & 402.1 (1.44$\times$) & \best{478.8 (1.72$\times$)} \\
 & SpeedBench      & 279.2 & 413.0 (1.48$\times$) & 419.6 (1.50$\times$) & 399.7 (1.43$\times$) & \best{494.6 (1.77$\times$)} \\
\midrule
\multicolumn{2}{l}{Average} & 278.8 & 844.2 (3.03$\times$) & 900.8 (3.24$\times$) & 838.0 (3.01$\times$) & \best{1042.4 (3.75$\times$)} \\
\bottomrule
\end{tabular}%
}
\end{minipage}\\[10pt]
\begin{minipage}{\linewidth}
\centering
\textbf{(b)} Sampling, $T{=}1$ (mean{\scriptsize$\pm$}std over seeds 42/43/44)\\[3pt]
\resizebox{\linewidth}{!}{%
\begin{tabular}{llccccc}
\toprule
Domain & Dataset & No-spec & DFlash & Domino & DSpark & \methodname \\
\midrule
\multirow{3}{*}{Math}
 & MATH-500       & \pmstd{277.8}{0.3} & \pmstd{429.1}{52.7} (1.55$\times$) & \pmstd{373.8}{98.2} (1.35$\times$) & \pmstd{343.2}{14.7} (1.24$\times$) & \best{\pmstd{545.5}{23.7} (1.96$\times$)} \\
 & AIME-25        & \pmstd{274.9}{0.6} & \pmstd{296.1}{81.6} (1.08$\times$) & \pmstd{253.1}{18.9} (0.92$\times$) & \pmstd{247.1}{9.3} (0.90$\times$) & \best{\pmstd{354.4}{27.7} (1.29$\times$)} \\
 & GSM8K          & \pmstd{277.7}{0.1} & \pmstd{610.7}{66.2} (2.20$\times$) & \best{\pmstd{692.5}{66.5} (2.49$\times$)} & \pmstd{569.3}{66.7} (2.05$\times$) & \pmstd{681.6}{34.1} (2.45$\times$) \\
\midrule
\multirow{3}{*}{Code}
 & HumanEval      & \pmstd{275.9}{0.0} & \pmstd{814.5}{21.0} (2.95$\times$) & \pmstd{883.3}{15.3} (3.20$\times$) & \pmstd{807.0}{31.2} (2.92$\times$) & \best{\pmstd{895.8}{67.9} (3.25$\times$)} \\
 & MBPP           & \pmstd{282.1}{0.0} & \pmstd{776.8}{2.2} (2.75$\times$) & \best{\pmstd{841.3}{12.8} (2.98$\times$)} & \pmstd{770.8}{7.9} (2.73$\times$) & \pmstd{793.4}{64.4} (2.81$\times$) \\
 & LiveCodeBench  & \pmstd{277.4}{0.1} & \best{\pmstd{643.3}{36.7} (2.32$\times$)} & \pmstd{619.9}{41.3} (2.24$\times$) & \pmstd{586.4}{45.6} (2.11$\times$) & \pmstd{545.3}{21.3} (1.97$\times$) \\
\midrule
\multirow{4}{*}{General}
 & MT-Bench       & \pmstd{281.4}{0.1} & \pmstd{364.3}{24.3} (1.30$\times$) & \pmstd{365.8}{5.3} (1.30$\times$) & \pmstd{346.3}{8.4} (1.23$\times$) & \best{\pmstd{442.3}{9.2} (1.57$\times$)} \\
 & Arena-Hard-v2  & \pmstd{277.8}{0.1} & \pmstd{224.7}{5.7} (0.81$\times$) & \pmstd{221.8}{1.0} (0.80$\times$) & \pmstd{221.8}{7.6} (0.80$\times$) & \best{\pmstd{310.7}{0.8} (1.12$\times$)} \\
 & Alpaca         & \pmstd{278.9}{0.1} & \pmstd{365.5}{3.9} (1.31$\times$) & \pmstd{370.0}{7.2} (1.33$\times$) & \pmstd{364.2}{6.2} (1.31$\times$) & \best{\pmstd{401.6}{5.0} (1.44$\times$)} \\
 & SpeedBench      & \pmstd{278.3}{0.1} & \pmstd{347.5}{5.0} (1.25$\times$) & \pmstd{351.1}{12.8} (1.26$\times$) & \pmstd{340.5}{17.4} (1.22$\times$) & \best{\pmstd{392.7}{5.5} (1.41$\times$)} \\
\midrule
\multicolumn{2}{l}{Average} & 278.2 & 487.2 (1.75$\times$) & 497.3 (1.79$\times$) & 459.6 (1.65$\times$) & \best{536.3 (1.93$\times$)} \\
\bottomrule
\end{tabular}%
}
\end{minipage}
\caption{\textbf{Throughput (output tokens/s, per-stream) and speedup
over no-spec decoding, Llama-3.1-8B, thinking off}, concurrency~1,
max 2048 generated tokens. Same format as Table~\ref{tab:main-q8b}.}
\label{tab:main-llama}
\end{table}

\begin{table}[h]
\centering
\small
\setlength{\tabcolsep}{3.5pt}
\resizebox{\linewidth}{!}{%
\begin{tabular}{llcccccccc}
\toprule
& & \multicolumn{2}{c}{DFlash} & \multicolumn{2}{c}{Domino} &
\multicolumn{2}{c}{DSpark} & \multicolumn{2}{c}{\methodname} \\
\cmidrule(lr){3-4}\cmidrule(lr){5-6}\cmidrule(lr){7-8}\cmidrule(lr){9-10}
Domain & Dataset & $T{=}0$ & $T{=}1$ & $T{=}0$ & $T{=}1$
 & $T{=}0$ & $T{=}1$ & $T{=}0$ & $T{=}1$ \\
\midrule
\multirow{3}{*}{Math}
 & MATH-500       & 7.194 & \pmstd{2.514}{0.075} & 7.837 & \pmstd{2.393}{0.278} & 7.540 & \pmstd{2.314}{0.067} & \best{8.141} & \best{\pmstd{2.635}{0.18}} \\
 & AIME-25        & 7.937 & \best{\pmstd{1.758}{0.149}} & 8.470 & \pmstd{1.662}{0.145} & 8.306 & \pmstd{1.573}{0.137} & \best{8.916} & \pmstd{1.714}{0.15} \\
 & GSM8K          & 4.529 & \pmstd{3.071}{0.374} & 4.692 & \best{\pmstd{3.496}{0.339}} & 4.543 & \pmstd{2.897}{0.242} & \best{5.984} & \pmstd{3.404}{0.25} \\
\midrule
\multirow{3}{*}{Code}
 & HumanEval      & 5.126 & \pmstd{4.186}{0.092} & 5.545 & \best{\pmstd{4.523}{0.059}} & 5.254 & \pmstd{4.258}{0.179} & \best{6.454} & \pmstd{4.490}{0.39} \\
 & MBPP           & 5.293 & \pmstd{3.929}{0.129} & \best{5.806} & \best{\pmstd{4.276}{0.091}} & 5.503 & \pmstd{4.000}{0.201} & 5.560 & \pmstd{4.274}{0.43} \\
 & LiveCodeBench  & 5.292 & \best{\pmstd{3.468}{0.176}} & 5.594 & \pmstd{3.283}{0.344} & 5.411 & \pmstd{3.142}{0.293} & \best{7.283} & \pmstd{2.649}{0.13} \\
\midrule
\multirow{4}{*}{General}
 & MT-Bench       & 2.546 & \pmstd{2.090}{0.120} & 2.644 & \best{\pmstd{2.155}{0.079}} & 2.593 & \pmstd{2.065}{0.056} & \best{2.802} & \pmstd{2.153}{0.01} \\
 & Arena-Hard-v2  & 2.466 & \pmstd{1.449}{0.024} & 2.457 & \pmstd{1.455}{0.003} & 2.401 & \pmstd{1.474}{0.029} & \best{2.773} & \best{\pmstd{1.499}{0.02}} \\
 & Alpaca         & 2.161 & \pmstd{1.872}{0.018} & 2.242 & \pmstd{1.924}{0.033} & 2.191 & \pmstd{1.950}{0.011} & \best{2.341} & \best{\pmstd{1.957}{0.0}} \\
 & SpeedBench      & 2.253 & \pmstd{1.837}{0.048} & 2.270 & \pmstd{1.852}{0.054} & 2.252 & \pmstd{1.857}{0.075} & \best{2.459} & \best{\pmstd{1.932}{0.031}} \\
\midrule
\multicolumn{2}{l}{Average} & 4.480 & 2.617 & 4.756 & \best{2.702} & 4.599 & 2.553 & \best{5.271} & 2.671 \\
\bottomrule
\end{tabular}%
}
\caption{\textbf{Mean accepted length $\tau$, Llama-3.1-8B, thinking
off}, companion to Table~\ref{tab:main-llama}. Domino attains
best $\tau$ on some datasets here, yet the lower
average throughput (Table~\ref{tab:main-llama}), the
$\tau$-vs-TPS trade at the aggregate level.}
\label{tab:tau-llama}
\end{table}

\begin{table}[h]
\centering
\small
\begin{minipage}{\linewidth}
\centering
\textbf{(a)} Greedy decoding, $T{=}0$ (single deterministic run)\\[3pt]
\resizebox{\linewidth}{!}{%
\begin{tabular}{llccccc}
\toprule
Domain & Dataset & No-spec & DFlash & Domino & DSpark & \methodname \\
\midrule
\multirow{3}{*}{Math}
 & MATH-500       & 343.2 & 1043.3 (3.04$\times$) & 1138.2 (3.32$\times$) & 1050.8 (3.06$\times$) & \best{1231.8 (3.59$\times$)} \\
 & AIME-25        & 342.9 & 928.1 (2.71$\times$) & 941.5 (2.75$\times$) & 906.7 (2.64$\times$) & \best{1074.5 (3.13$\times$)} \\
 & GSM8K          & 348.0 & 733.5 (2.11$\times$) & 728.3 (2.09$\times$) & 698.2 (2.01$\times$) & \best{854.8 (2.46$\times$)} \\
\midrule
\multirow{3}{*}{Code}
 & HumanEval      & 335.2 & 874.8 (2.61$\times$) & 880.9 (2.63$\times$) & 847.7 (2.53$\times$) & \best{1003.8 (3.00$\times$)} \\
 & MBPP           & 333.5 & 759.5 (2.28$\times$) & 779.0 (2.34$\times$) & 734.6 (2.20$\times$) & \best{974.0 (2.92$\times$)} \\
 & LiveCodeBench  & 338.9 & 684.5 (2.02$\times$) & 743.9 (2.20$\times$) & 698.9 (2.06$\times$) & \best{890.8 (2.63$\times$)} \\
\midrule
\multirow{4}{*}{General}
 & MT-Bench       & 337.0 & 317.6 (0.94$\times$) & 316.9 (0.94$\times$) & 298.8 (0.89$\times$) & \best{510.1 (1.51$\times$)} \\
 & Arena-Hard-v2  & 333.5 & 357.1 (1.07$\times$) & 347.0 (1.04$\times$) & 336.2 (1.01$\times$) & \best{456.5 (1.37$\times$)} \\
 & Alpaca         & 348.0 & 298.3 (0.86$\times$) & 302.7 (0.87$\times$) & 296.9 (0.85$\times$) & \best{382.7 (1.10$\times$)} \\
 & SpeedBench      & 341.5 & 427.2 (1.25$\times$) & 422.8 (1.24$\times$) & 403.7 (1.18$\times$) & \best{472.0 (1.38$\times$)} \\
\midrule
\multicolumn{2}{l}{Average} & 340.2 & 642.4 (1.89$\times$) & 660.1 (1.94$\times$) & 627.2 (1.84$\times$) & \best{785.1 (2.31$\times$)} \\
\bottomrule
\end{tabular}%
}
\end{minipage}\\[10pt]
\begin{minipage}{\linewidth}
\centering
\textbf{(b)} Sampling, $T{=}1$ (mean{\scriptsize$\pm$}std over seeds 42/43/44)\\[3pt]
\resizebox{\linewidth}{!}{%
\begin{tabular}{llccccc}
\toprule
Domain & Dataset & No-spec & DFlash & Domino & DSpark & \methodname \\
\midrule
\multirow{3}{*}{Math}
 & MATH-500       & \pmstd{337.5}{0.2} & \pmstd{923.6}{17.5} (2.74$\times$) & \pmstd{945.3}{16.2} (2.80$\times$) & \pmstd{902.2}{8.7} (2.67$\times$) & \best{\pmstd{1060.9}{2.7} (3.14$\times$)} \\
 & AIME-25        & \pmstd{337.2}{0.1} & \pmstd{646.8}{59.5} (1.92$\times$) & \pmstd{666.5}{67.1} (1.98$\times$) & \pmstd{632.3}{48.4} (1.88$\times$) & \best{\pmstd{808.8}{19.1} (2.40$\times$)} \\
 & GSM8K          & \pmstd{342.1}{0.2} & \pmstd{698.8}{15.6} (2.04$\times$) & \pmstd{687.1}{13.3} (2.01$\times$) & \pmstd{676.2}{4.7} (1.98$\times$) & \best{\pmstd{793.8}{8.5} (2.32$\times$)} \\
\midrule
\multirow{3}{*}{Code}
 & HumanEval      & \pmstd{329.5}{0.1} & \pmstd{827.3}{25.7} (2.51$\times$) & \pmstd{816.0}{3.5} (2.48$\times$) & \pmstd{782.2}{16.1} (2.37$\times$) & \best{\pmstd{908.3}{14.9} (2.76$\times$)} \\
 & MBPP           & \pmstd{327.8}{0.1} & \pmstd{728.1}{5.5} (2.22$\times$) & \pmstd{749.8}{12.0} (2.29$\times$) & \pmstd{708.8}{4.6} (2.16$\times$) & \best{\pmstd{812.2}{9.6} (2.48$\times$)} \\
 & LiveCodeBench  & \pmstd{333.1}{0.3} & \pmstd{440.2}{26.2} (1.32$\times$) & \pmstd{446.6}{10.4} (1.34$\times$) & \pmstd{444.7}{10.4} (1.34$\times$) & \best{\pmstd{559.5}{4.8} (1.68$\times$)} \\
\midrule
\multirow{4}{*}{General}
 & MT-Bench       & \pmstd{331.0}{0.2} & \pmstd{308.0}{8.6} (0.93$\times$) & \pmstd{313.2}{13.2} (0.95$\times$) & \pmstd{301.9}{5.8} (0.91$\times$) & \best{\pmstd{475.8}{10.3} (1.44$\times$)} \\
 & Arena-Hard-v2  & \pmstd{327.9}{0.0} & \pmstd{331.0}{8.1} (1.01$\times$) & \pmstd{290.9}{0.9} (0.89$\times$) & \pmstd{298.2}{10.4} (0.91$\times$) & \best{\pmstd{386.1}{2.7} (1.18$\times$)} \\
 & Alpaca         & \pmstd{342.1}{0.2} & \pmstd{289.9}{6.2} (0.85$\times$) & \pmstd{296.5}{6.9} (0.87$\times$) & \pmstd{291.4}{1.1} (0.85$\times$) & \best{\pmstd{377.3}{1.8} (1.10$\times$)} \\
 & SpeedBench      & \pmstd{336.3}{0.9} & \pmstd{375.3}{44.2} (1.12$\times$) & \pmstd{365.7}{43.5} (1.09$\times$) & \pmstd{362.0}{43.0} (1.08$\times$) & \best{\pmstd{440.1}{29.5} (1.31$\times$)} \\
\midrule
\multicolumn{2}{l}{Average} & 334.4 & 556.9 (1.67$\times$) & 557.8 (1.67$\times$) & 540.0 (1.62$\times$) & \best{662.3 (1.98$\times$)} \\
\bottomrule
\end{tabular}%
}
\end{minipage}
\caption{\textbf{Throughput (output tokens/s, per-stream) and speedup
over no-spec decoding, Qwen3-Coder-30B-A3B, thinking off}, concurrency~1,
max 2048 generated tokens. Same format as Table~\ref{tab:main-q8b}.}
\label{tab:main-q30b}
\end{table}

\begin{table}[h]
\centering
\small
\setlength{\tabcolsep}{3.5pt}
\resizebox{\linewidth}{!}{%
\begin{tabular}{llcccccccc}
\toprule
& & \multicolumn{2}{c}{DFlash} & \multicolumn{2}{c}{Domino} &
\multicolumn{2}{c}{DSpark} & \multicolumn{2}{c}{\methodname} \\
\cmidrule(lr){3-4}\cmidrule(lr){5-6}\cmidrule(lr){7-8}\cmidrule(lr){9-10}
Domain & Dataset & $T{=}0$ & $T{=}1$ & $T{=}0$ & $T{=}1$
 & $T{=}0$ & $T{=}1$ & $T{=}0$ & $T{=}1$ \\
\midrule
\multirow{3}{*}{Math}
 & MATH-500       & 6.372 & \pmstd{5.639}{0.062} & 6.947 & \pmstd{6.017}{0.104} & 6.536 & \pmstd{5.626}{0.055} & \best{7.224} & \best{\pmstd{6.221}{0.016}} \\
 & AIME-25        & 5.566 & \pmstd{4.276}{0.122} & 6.057 & \pmstd{4.439}{0.091} & 5.695 & \pmstd{4.309}{0.091} & \best{6.312} & \best{\pmstd{4.693}{0.114}} \\
 & GSM8K          & 4.717 & \pmstd{4.457}{0.138} & 4.998 & \pmstd{4.616}{0.109} & 4.652 & \pmstd{4.372}{0.029} & \best{5.322} & \best{\pmstd{5.036}{0.064}} \\
\midrule
\multirow{3}{*}{Code}
 & HumanEval      & 5.805 & \pmstd{5.279}{0.155} & 6.189 & \pmstd{5.598}{0.031} & 5.789 & \pmstd{5.285}{0.044} & \best{6.435} & \best{\pmstd{5.803}{0.103}} \\
 & MBPP           & 4.981 & \pmstd{4.787}{0.023} & 5.391 & \best{\pmstd{5.208}{0.048}} & 4.989 & \pmstd{4.828}{0.029} & \best{6.141} & \pmstd{4.987}{0.063} \\
 & LiveCodeBench  & 4.498 & \pmstd{3.200}{0.045} & 4.985 & \pmstd{3.279}{0.079} & 4.668 & \pmstd{3.221}{0.076} & \best{5.441} & \best{\pmstd{3.331}{0.032}} \\
\midrule
\multirow{4}{*}{General}
 & MT-Bench       & 2.679 & \pmstd{2.516}{0.024} & 2.807 & \pmstd{2.674}{0.054} & 2.669 & \pmstd{2.531}{0.038} & \best{2.997} & \best{\pmstd{2.762}{0.069}} \\
 & Arena-Hard-v2  & 2.476 & \pmstd{2.124}{0.026} & 2.524 & \pmstd{2.143}{0.020} & 2.400 & \pmstd{2.073}{0.016} & \best{2.648} & \best{\pmstd{2.205}{0.018}} \\
 & Alpaca         & 2.115 & \pmstd{2.028}{0.049} & 2.228 & \pmstd{2.112}{0.030} & 2.119 & \pmstd{2.044}{0.049} & \best{2.234} & \best{\pmstd{2.182}{0.010}} \\
 & SpeedBench      & 2.704 & \pmstd{2.393}{0.299} & 2.837 & \pmstd{2.458}{0.298} & 2.640 & \pmstd{2.404}{0.268} & \best{2.989} & \best{\pmstd{2.762}{0.221}} \\
\midrule
\multicolumn{2}{l}{Average} & 4.191 & 3.670 & 4.496 & 3.854 & 4.216 & 3.669 & \best{4.774} & \best{3.998} \\
\bottomrule
\end{tabular}%
}
\caption{\textbf{Mean accepted length $\tau$, Qwen3-Coder-30B-A3B, thinking off},
companion to Table~\ref{tab:main-q30b}.}
\label{tab:tau-q30b}
\end{table}

\begin{table}[h]
\centering
\small
\begin{minipage}{\linewidth}
\centering
\textbf{(a)} Greedy decoding, $T{=}0$ (single deterministic run)\\[3pt]
\resizebox{\linewidth}{!}{%
\begin{tabular}{llccccc}
\toprule
Domain & Dataset & No-spec & DFlash & Domino & DSpark & \methodname \\
\midrule
\multirow{3}{*}{Math}
 & MATH-500       & 366.2 & 1764.7 (4.82$\times$) & 1974.6 (5.39$\times$) & 1599.5 (4.37$\times$) & \best{2165.5 (5.91$\times$)} \\
 & AIME-25        & 364.4 & 1640.5 (4.50$\times$) & 1922.1 (5.27$\times$) & 1566.0 (4.30$\times$) & \best{2085.9 (5.72$\times$)} \\
 & GSM8K          & 369.6 & 1330.9 (3.60$\times$) & 1450.5 (3.93$\times$) & 1127.0 (3.05$\times$) & \best{1638.4 (4.43$\times$)} \\
\midrule
\multirow{3}{*}{Code}
 & HumanEval      & 370.6 & 1226.4 (3.31$\times$) & 1348.7 (3.64$\times$) & 1075.4 (2.90$\times$) & \best{1475.8 (3.98$\times$)} \\
 & MBPP           & 370.2 & 1137.6 (3.07$\times$) & 1267.5 (3.42$\times$) & 978.6 (2.64$\times$) & \best{1435.3 (3.88$\times$)} \\
 & LiveCodeBench  & 369.2 & 1482.9 (4.02$\times$) & 1620.5 (4.39$\times$) & 1309.8 (3.55$\times$) & \best{1754.1 (4.75$\times$)} \\
\midrule
\multirow{4}{*}{General}
 & MT-Bench       & 369.0 & 580.4 (1.57$\times$) & 593.4 (1.61$\times$) & 528.2 (1.43$\times$) & \best{661.3 (1.79$\times$)} \\
 & Arena-Hard-v2  & 369.6 & 600.8 (1.63$\times$) & 610.5 (1.65$\times$) & 527.7 (1.43$\times$) & \best{692.3 (1.87$\times$)} \\
 & Alpaca         & 368.1 & 559.0 (1.52$\times$) & 595.6 (1.62$\times$) & 513.0 (1.39$\times$) & \best{658.7 (1.79$\times$)} \\
 & SpeedBench      & 369.1 & 645.6 (1.75$\times$) & 643.9 (1.74$\times$) & 575.5 (1.56$\times$) & \best{759.0 (2.06$\times$)} \\
\midrule
\multicolumn{2}{l}{Average} & 368.6 & 1096.9 (2.98$\times$) & 1202.7 (3.27$\times$) & 980.1 (2.66$\times$) & \best{1332.6 (3.62$\times$)} \\
\bottomrule
\end{tabular}%
}
\end{minipage}\\[10pt]
\begin{minipage}{\linewidth}
\centering
\textbf{(b)} Sampling, $T{=}1$ (mean{\scriptsize$\pm$}std over seeds 42/43/44)\\[3pt]
\resizebox{\linewidth}{!}{%
\begin{tabular}{llccccc}
\toprule
Domain & Dataset & No-spec & DFlash & Domino & DSpark & \methodname \\
\midrule
\multirow{3}{*}{Math}
 & MATH-500       & \pmstd{365.2}{0.0} & \pmstd{1360.4}{42.2} (3.73$\times$) & \pmstd{1483.8}{47.3} (4.06$\times$) & \pmstd{1277.3}{33.5} (3.50$\times$) & \best{\pmstd{1644.7}{21.0} (4.50$\times$)} \\
 & AIME-25        & \pmstd{364.0}{0.1} & \pmstd{1017.5}{78.9} (2.80$\times$) & \pmstd{1149.8}{18.1} (3.16$\times$) & \pmstd{966.8}{44.1} (2.66$\times$) & \best{\pmstd{1194.1}{142.7} (3.28$\times$)} \\
 & GSM8K          & \pmstd{368.6}{0.1} & \pmstd{1270.6}{14.1} (3.45$\times$) & \pmstd{1375.3}{25.2} (3.73$\times$) & \pmstd{1080.4}{10.8} (2.93$\times$) & \best{\pmstd{1525.7}{19.7} (4.14$\times$)} \\
\midrule
\multirow{3}{*}{Code}
 & HumanEval      & \pmstd{370.1}{0.0} & \pmstd{1150.9}{10.6} (3.11$\times$) & \pmstd{1231.8}{10.8} (3.33$\times$) & \pmstd{1026.9}{18.0} (2.77$\times$) & \best{\pmstd{1372.1}{14.1} (3.71$\times$)} \\
 & MBPP           & \pmstd{369.4}{0.1} & \pmstd{1032.1}{26.3} (2.79$\times$) & \pmstd{1133.5}{35.7} (3.07$\times$) & \pmstd{928.2}{14.9} (2.51$\times$) & \best{\pmstd{1260.7}{15.5} (3.41$\times$)} \\
 & LiveCodeBench  & \pmstd{369.2}{0.8} & \pmstd{1184.6}{22.8} (3.21$\times$) & \best{\pmstd{1302.5}{29.7} (3.53$\times$)} & \pmstd{1065.5}{32.3} (2.89$\times$) & \pmstd{1301.1}{39.4} (3.52$\times$) \\
\midrule
\multirow{4}{*}{General}
 & MT-Bench       & \pmstd{369.0}{0.1} & \pmstd{559.4}{3.7} (1.52$\times$) & \pmstd{570.1}{5.1} (1.55$\times$) & \pmstd{512.1}{7.9} (1.39$\times$) & \best{\pmstd{639.1}{17.1} (1.73$\times$)} \\
 & Arena-Hard-v2  & \pmstd{369.9}{0.4} & \pmstd{568.5}{12.0} (1.54$\times$) & \pmstd{554.5}{9.5} (1.50$\times$) & \pmstd{490.8}{8.7} (1.33$\times$) & \best{\pmstd{631.4}{25.0} (1.71$\times$)} \\
 & Alpaca         & \pmstd{367.2}{0.0} & \pmstd{548.3}{6.8} (1.49$\times$) & \pmstd{572.1}{7.0} (1.56$\times$) & \pmstd{517.3}{3.0} (1.41$\times$) & \best{\pmstd{631.3}{8.6} (1.72$\times$)} \\
 & SpeedBench      & \pmstd{368.8}{0.2} & \pmstd{583.4}{11.6} (1.58$\times$) & \pmstd{583.4}{6.5} (1.58$\times$) & \pmstd{518.6}{11.4} (1.41$\times$) & \best{\pmstd{668.3}{21.9} (1.81$\times$)} \\
\midrule
\multicolumn{2}{l}{Average} & 368.1 & 927.6 (2.52$\times$) & 995.7 (2.71$\times$) & 838.4 (2.28$\times$) & \best{1086.8 (2.95$\times$)} \\
\bottomrule
\end{tabular}%
}
\end{minipage}
\caption{\textbf{Throughput (output tokens/s, per-stream) and speedup
over no-spec decoding, Qwen3-4B, thinking off}, concurrency~1, max
2048 generated tokens. Same format as Table~\ref{tab:main-q8b}.}
\label{tab:main-q4b}
\end{table}

\begin{table}[h]
\centering
\small
\setlength{\tabcolsep}{3.5pt}
\resizebox{\linewidth}{!}{%
\begin{tabular}{llcccccccc}
\toprule
& & \multicolumn{2}{c}{DFlash} & \multicolumn{2}{c}{Domino} &
\multicolumn{2}{c}{DSpark} & \multicolumn{2}{c}{\methodname} \\
\cmidrule(lr){3-4}\cmidrule(lr){5-6}\cmidrule(lr){7-8}\cmidrule(lr){9-10}
Domain & Dataset & $T{=}0$ & $T{=}1$ & $T{=}0$ & $T{=}1$
 & $T{=}0$ & $T{=}1$ & $T{=}0$ & $T{=}1$ \\
\midrule
\multirow{3}{*}{Math}
 & MATH-500       & 6.443 & \pmstd{5.104}{0.039} & 7.719 & \pmstd{5.799}{0.119} & 6.125 & \pmstd{4.910}{0.068} & \best{7.967} & \best{\pmstd{6.041}{0.098}} \\
 & AIME-25        & 5.996 & \pmstd{3.894}{0.084} & 7.176 & \pmstd{4.520}{0.067} & 5.729 & \pmstd{3.872}{0.068} & \best{7.368} & \best{\pmstd{4.623}{0.178}} \\
 & GSM8K          & 4.872 & \pmstd{4.604}{0.036} & 5.825 & \pmstd{5.422}{0.084} & 4.474 & \pmstd{4.230}{0.021} & \best{6.082} & \best{\pmstd{5.610}{0.065}} \\
\midrule
\multirow{3}{*}{Code}
 & HumanEval      & 4.416 & \pmstd{4.163}{0.012} & 5.263 & \pmstd{4.904}{0.034} & 4.091 & \pmstd{3.848}{0.021} & \best{5.351} & \best{\pmstd{4.949}{0.046}} \\
 & MBPP           & 4.270 & \pmstd{3.812}{0.084} & 5.085 & \pmstd{4.508}{0.062} & 3.933 & \pmstd{3.541}{0.004} & \best{5.297} & \best{\pmstd{4.603}{0.007}} \\
 & LiveCodeBench  & 6.075 & \pmstd{4.444}{0.053} & 7.103 & \best{\pmstd{5.178}{0.044}} & 5.637 & \pmstd{4.143}{0.076} & \best{7.127} & \pmstd{5.110}{0.139} \\
\midrule
\multirow{4}{*}{General}
 & MT-Bench       & 2.564 & \pmstd{2.351}{0.017} & 2.851 & \pmstd{2.592}{0.010} & 2.415 & \pmstd{2.250}{0.023} & \best{2.946} & \best{\pmstd{2.669}{0.033}} \\
 & Arena-Hard-v2  & 2.408 & \pmstd{2.173}{0.036} & 2.596 & \pmstd{2.278}{0.005} & 2.200 & \pmstd{1.980}{0.011} & \best{2.716} & \best{\pmstd{2.393}{0.036}} \\
 & Alpaca         & 2.199 & \pmstd{2.071}{0.012} & 2.420 & \pmstd{2.278}{0.032} & 2.086 & \pmstd{2.026}{0.029} & \best{2.515} & \best{\pmstd{2.357}{0.018}} \\
 & SpeedBench      & 2.381 & \pmstd{2.143}{0.029} & 2.566 & \pmstd{2.290}{0.030} & 2.200 & \pmstd{2.005}{0.031} & \best{2.753} & \best{\pmstd{2.425}{0.052}} \\
\midrule
\multicolumn{2}{l}{Average} & 4.162 & 3.476 & 4.861 & 3.977 & 3.889 & 3.281 & \best{5.012} & \best{4.078} \\
\bottomrule
\end{tabular}%
}
\caption{\textbf{Mean accepted length $\tau$, Qwen3-4B, thinking off},
companion to Table~\ref{tab:main-q4b}. \methodname leads $\tau$ in 19 of
20 cells and both averages.}
\label{tab:tau-q4b}
\end{table}

\begin{table}[h]
\centering
\small
\begin{minipage}{\linewidth}
\centering
\textbf{(a)} Greedy decoding, $T{=}0$ (single deterministic run)\\[3pt]
\resizebox{\linewidth}{!}{%
\begin{tabular}{llccccc}
\toprule
Domain & Dataset & No-spec & DFlash & Domino & DSpark & \methodname \\
\midrule
\multirow{3}{*}{Math}
 & MATH-500       & 361.0 & 868.8 (2.41$\times$) & 921.7 (2.55$\times$) & 785.8 (2.18$\times$) & \best{1000.4 (2.77$\times$)} \\
 & AIME-25        & 359.5 & 861.5 (2.40$\times$) & 909.9 (2.53$\times$) & 755.0 (2.10$\times$) & \best{989.8 (2.75$\times$)} \\
 & GSM8K          & 362.0 & 846.8 (2.34$\times$) & 869.7 (2.40$\times$) & 735.9 (2.03$\times$) & \best{974.0 (2.69$\times$)} \\
\midrule
\multirow{3}{*}{Code}
 & HumanEval      & 357.9 & 824.4 (2.30$\times$) & 875.9 (2.45$\times$) & 739.7 (2.07$\times$) & \best{934.4 (2.61$\times$)} \\
 & MBPP           & 359.5 & 797.0 (2.22$\times$) & 836.3 (2.33$\times$) & 731.1 (2.03$\times$) & \best{923.9 (2.57$\times$)} \\
 & LiveCodeBench  & 361.2 & 780.1 (2.16$\times$) & 842.8 (2.33$\times$) & 706.5 (1.96$\times$) & \best{896.5 (2.48$\times$)} \\
\midrule
\multirow{4}{*}{General}
 & MT-Bench       & 363.4 & 741.7 (2.04$\times$) & 745.6 (2.05$\times$) & 651.0 (1.79$\times$) & \best{839.0 (2.31$\times$)} \\
 & Arena-Hard-v2  & 362.7 & 682.9 (1.88$\times$) & 702.1 (1.94$\times$) & 574.1 (1.58$\times$) & \best{796.0 (2.20$\times$)} \\
 & Alpaca         & 367.9 & 543.2 (1.48$\times$) & 551.2 (1.50$\times$) & 497.4 (1.35$\times$) & \best{604.8 (1.64$\times$)} \\
 & SpeedBench      & 365.7 & 588.7 (1.61$\times$) & 596.9 (1.63$\times$) & 529.2 (1.45$\times$) & \best{663.5 (1.81$\times$)} \\
\midrule
\multicolumn{2}{l}{Average} & 362.1 & 753.5 (2.08$\times$) & 785.2 (2.17$\times$) & 670.6 (1.85$\times$) & \best{862.2 (2.38$\times$)} \\
\bottomrule
\end{tabular}%
}
\end{minipage}\\[10pt]
\begin{minipage}{\linewidth}
\centering
\textbf{(b)} Sampling, $T{=}1$ (mean{\scriptsize$\pm$}std over seeds 42/43/44)\\[3pt]
\resizebox{\linewidth}{!}{%
\begin{tabular}{llccccc}
\toprule
Domain & Dataset & No-spec & DFlash & Domino & DSpark & \methodname \\
\midrule
\multirow{3}{*}{Math}
 & MATH-500       & \pmstd{360.6}{0.0} & \pmstd{812.2}{1.5} (2.25$\times$) & \pmstd{861.0}{10.2} (2.39$\times$) & \pmstd{731.0}{4.3} (2.03$\times$) & \best{\pmstd{915.0}{9.1} (2.54$\times$)} \\
 & AIME-25        & \pmstd{359.1}{0.0} & \pmstd{775.0}{26.0} (2.16$\times$) & \pmstd{799.8}{10.0} (2.23$\times$) & \pmstd{692.2}{4.3} (1.93$\times$) & \best{\pmstd{905.8}{20.8} (2.52$\times$)} \\
 & GSM8K          & \pmstd{360.5}{0.2} & \pmstd{756.2}{3.0} (2.10$\times$) & \pmstd{777.3}{6.0} (2.16$\times$) & \pmstd{669.0}{2.3} (1.86$\times$) & \best{\pmstd{859.6}{5.0} (2.38$\times$)} \\
\midrule
\multirow{3}{*}{Code}
 & HumanEval      & \pmstd{357.9}{0.4} & \pmstd{735.6}{2.8} (2.06$\times$) & \pmstd{775.1}{4.7} (2.17$\times$) & \pmstd{672.1}{2.9} (1.88$\times$) & \best{\pmstd{832.5}{4.3} (2.33$\times$)} \\
 & MBPP           & \pmstd{358.4}{0.6} & \pmstd{697.7}{2.5} (1.95$\times$) & \pmstd{739.1}{2.0} (2.06$\times$) & \pmstd{647.6}{2.4} (1.81$\times$) & \best{\pmstd{804.9}{4.3} (2.25$\times$)} \\
 & LiveCodeBench  & \pmstd{360.7}{0.0} & \pmstd{689.4}{4.7} (1.91$\times$) & \pmstd{726.2}{6.0} (2.01$\times$) & \pmstd{632.6}{1.4} (1.75$\times$) & \best{\pmstd{778.1}{8.7} (2.16$\times$)} \\
\midrule
\multirow{4}{*}{General}
 & MT-Bench       & \pmstd{361.2}{0.5} & \pmstd{622.4}{9.1} (1.72$\times$) & \pmstd{637.3}{4.2} (1.76$\times$) & \pmstd{566.7}{8.4} (1.57$\times$) & \best{\pmstd{697.9}{14.5} (1.93$\times$)} \\
 & Arena-Hard-v2  & \pmstd{362.5}{0.1} & \pmstd{584.5}{6.4} (1.61$\times$) & \pmstd{579.2}{3.6} (1.60$\times$) & \pmstd{519.6}{4.8} (1.43$\times$) & \best{\pmstd{657.2}{6.8} (1.81$\times$)} \\
 & Alpaca         & \pmstd{367.7}{0.2} & \pmstd{512.7}{2.0} (1.39$\times$) & \pmstd{527.6}{3.1} (1.44$\times$) & \pmstd{477.3}{1.8} (1.30$\times$) & \best{\pmstd{576.8}{2.6} (1.57$\times$)} \\
 & SpeedBench      & \pmstd{364.4}{0.2} & \pmstd{549.3}{9.5} (1.51$\times$) & \pmstd{544.7}{3.1} (1.50$\times$) & \pmstd{493.6}{2.2} (1.36$\times$) & \best{\pmstd{625.2}{5.4} (1.72$\times$)} \\
\midrule
\multicolumn{2}{l}{Average} & 361.3 & 673.5 (1.87$\times$) & 696.7 (1.93$\times$) & 610.2 (1.69$\times$) & \best{765.3 (2.12$\times$)} \\
\bottomrule
\end{tabular}%
}
\end{minipage}
\caption{\textbf{Throughput (output tokens/s, per-stream) and speedup
over no-spec decoding, Qwen3-4B, thinking on}, concurrency~1, max
4096 generated tokens. Same format as Table~\ref{tab:main-q8b}.}
\label{tab:main-q4b-think}
\end{table}

\begin{table}[h]
\centering
\small
\setlength{\tabcolsep}{3.5pt}
\resizebox{\linewidth}{!}{%
\begin{tabular}{llcccccccc}
\toprule
& & \multicolumn{2}{c}{DFlash} & \multicolumn{2}{c}{Domino} &
\multicolumn{2}{c}{DSpark} & \multicolumn{2}{c}{\methodname} \\
\cmidrule(lr){3-4}\cmidrule(lr){5-6}\cmidrule(lr){7-8}\cmidrule(lr){9-10}
Domain & Dataset & $T{=}0$ & $T{=}1$ & $T{=}0$ & $T{=}1$
 & $T{=}0$ & $T{=}1$ & $T{=}0$ & $T{=}1$ \\
\midrule
\multirow{3}{*}{Math}
 & MATH-500       & 3.216 & \pmstd{2.957}{0.008} & 3.607 & \pmstd{3.289}{0.005} & 2.969 & \pmstd{2.771}{0.001} & \best{3.644} & \best{\pmstd{3.317}{0.012}} \\
 & AIME-25        & 3.140 & \pmstd{2.845}{0.032} & 3.520 & \pmstd{3.134}{0.026} & 2.900 & \pmstd{2.664}{0.017} & \best{3.568} & \best{\pmstd{3.234}{0.020}} \\
 & GSM8K          & 3.099 & \pmstd{2.753}{0.008} & 3.437 & \pmstd{3.030}{0.010} & 2.848 & \pmstd{2.563}{0.018} & \best{3.538} & \best{\pmstd{3.088}{0.010}} \\
\midrule
\multirow{3}{*}{Code}
 & HumanEval      & 3.100 & \pmstd{2.689}{0.010} & \best{3.517} & \best{\pmstd{3.020}{0.020}} & 2.858 & \pmstd{2.524}{0.012} & 3.451 & \pmstd{2.968}{0.002} \\
 & MBPP           & 2.961 & \pmstd{2.564}{0.004} & \best{3.337} & \best{\pmstd{2.850}{0.005}} & 2.749 & \pmstd{2.425}{0.012} & 3.327 & \pmstd{2.847}{0.000} \\
 & LiveCodeBench  & 2.880 & \pmstd{2.553}{0.019} & \best{3.267} & \best{\pmstd{2.830}{0.014}} & 2.674 & \pmstd{2.403}{0.012} & 3.246 & \pmstd{2.816}{0.012} \\
\midrule
\multirow{4}{*}{General}
 & MT-Bench       & 2.531 & \pmstd{2.246}{0.014} & 2.784 & \pmstd{2.440}{0.018} & 2.356 & \pmstd{2.118}{0.006} & \best{2.843} & \best{\pmstd{2.471}{0.004}} \\
 & Arena-Hard-v2  & 2.482 & \pmstd{2.064}{0.016} & 2.695 & \pmstd{2.223}{0.007} & 2.230 & \pmstd{1.934}{0.003} & \best{2.807} & \best{\pmstd{2.279}{0.008}} \\
 & Alpaca         & 2.148 & \pmstd{1.921}{0.007} & 2.333 & \pmstd{2.075}{0.013} & 2.013 & \pmstd{1.841}{0.006} & \best{2.385} & \best{\pmstd{2.134}{0.023}} \\
 & SpeedBench      & 2.225 & \pmstd{2.010}{0.026} & 2.393 & \pmstd{2.132}{0.013} & 2.053 & \pmstd{1.886}{0.007} & \best{2.478} & \best{\pmstd{2.256}{0.007}} \\
\midrule
\multicolumn{2}{l}{Average} & 2.778 & 2.460 & 3.089 & 2.702 & 2.565 & 2.313 & \best{3.129} & \best{2.741} \\
\bottomrule
\end{tabular}%
}
\caption{\textbf{Mean accepted length $\tau$, Qwen3-4B, thinking on},
companion to Table~\ref{tab:main-q4b-think}.}
\label{tab:tau-q4b-think}
\end{table}

\clearpage

\section{Coverage by verifier}
\label{app:coverage}
 
Table~\ref{tab:coverage} summarises Figure~\ref{fig:coverage} for all
four verifiers, and the pattern is strikingly uniform: at the first
drafted slot the top-16 pool contains the committed token about
$95\%$ of the time and the drafter's own first choice about $81\%$,
while by the last slot the pool still holds it in $57$--$61\%$ of
cases but the first choice finds it only about one time in five.
In other words, deep in the block roughly every second slot still has
the right token waiting in the candidate pool, the drafter simply
stops picking it. This is the headroom the corrector re-selects from.
 
\begin{table}[h]
\centering
\small
\begin{tabular}{lcccc}
\toprule
 & top-1 & top-4 & top-8 & top-16 \\
\midrule
Qwen3-8B, mean over drafted slots       & 0.400 & 0.587 & 0.676 & 0.755 \\
\quad first drafted slot                & 0.814 & 0.915 & 0.936 & 0.951 \\
\quad last drafted slot                 & 0.154 & 0.355 & 0.483 & 0.608 \\
\midrule
Qwen3-4B, mean over drafted slots       & 0.413 & 0.574 & 0.651 & 0.722 \\
\quad first drafted slot                & 0.815 & 0.907 & 0.928 & 0.943 \\
\quad last drafted slot                 & 0.169 & 0.348 & 0.461 & 0.573 \\
\midrule
Qwen3-Coder-30B-A3B, mean over drafted slots & 0.423 & 0.584 & 0.657 & 0.723 \\
\quad first drafted slot                & 0.811 & 0.900 & 0.920 & 0.936 \\
\quad last drafted slot                 & 0.181 & 0.366 & 0.476 & 0.582 \\
\midrule
Llama-3.1-8B, mean over drafted slots   & 0.449 & 0.602 & 0.672 & 0.735 \\
\quad first drafted slot                & 0.830 & 0.915 & 0.933 & 0.946 \\
\quad last drafted slot                 & 0.205 & 0.377 & 0.482 & 0.585 \\
\bottomrule
\end{tabular}
\caption{\textbf{Coverage by verifier}: means over the 15 drafted
slots of Figure~\ref{fig:coverage}, with the first and last slot shown
per verifier. Top-1 is the drafter's own per-slot accuracy;
\methodname{} uses $K{=}16$.}
\label{tab:coverage}
\end{table}
\clearpage

\section{Throughput at higher concurrency}
\label{app:concurrency}

\begin{figure}[t]
    \centering
    \includegraphics[width=\linewidth]{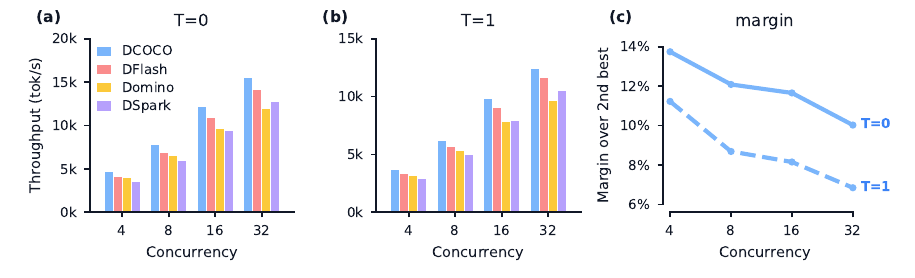}
    \caption{\textbf{At high concurrency, \methodname achieves the highest throughput, while Domino and DSpark fall below DFlash.}
    Qwen3-4B throughput at (a) $T=0$ and (b) $T=1$, averaged over the ten suites.
    (c) \methodname's relative throughput gain over the strongest
    baseline at each concurrency. \methodname's margin is $10$--$14\%$ at $T{=}0$ and $7$--$11\%$ at $T{=}1$.}
    \label{fig:q4b_conc}
\end{figure}

\begin{figure}[t]
    \centering
    \includegraphics[width=\linewidth]{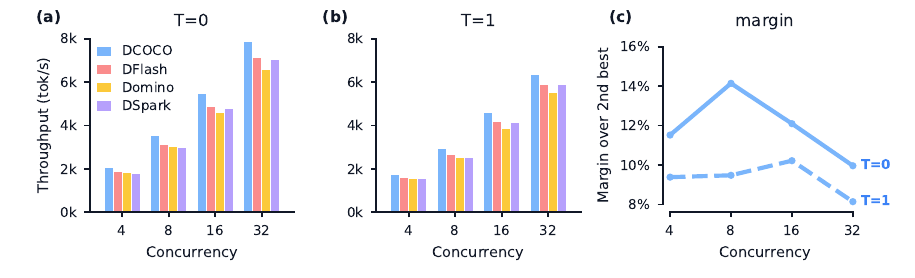}
    \caption{\textbf{\methodname achieves the highest throughput at every concurrency, while Domino and DSpark stay at or below DFlash.}
    Qwen3-Coder-30B-A3B throughput at (a) $T=0$ and (b) $T=1$, averaged over the ten suites.
    (c) \methodname's relative throughput gain over the strongest
    baseline at each concurrency. \methodname's margin is $10$--$14\%$ at $T{=}0$ and $8$--$10\%$ at $T{=}1$.}
    \label{fig:q30b_conc}
\end{figure}

\begin{figure}[t]
    \centering
    \includegraphics[width=\linewidth]{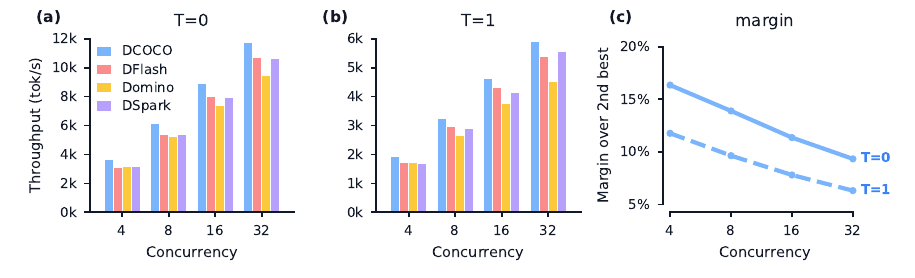}
    \caption{\textbf{\methodname achieves the highest throughput at every concurrency, while Domino falls increasingly behind DFlash as load grows.}
    Llama-3.1-8B throughput at (a) $T=0$ and (b) $T=1$, averaged over the ten suites.
    (c) \methodname's relative throughput gain over the strongest
    baseline at each concurrency. \methodname's margin is $9$--$16\%$ at $T{=}0$ and $6$--$12\%$ at $T{=}1$.}
    \label{fig:llama_conc}
\end{figure}

\begin{figure}[t]
    \centering
    \includegraphics[width=\linewidth]{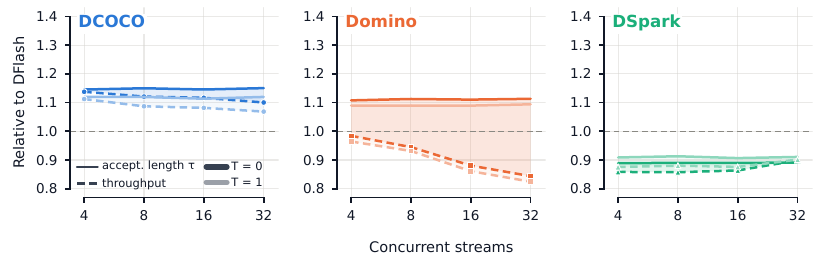}

    \caption{\textbf{Domino's acceptance gain is consumed by per-cycle overhead as concurrency grows, and DSpark never matches DFlash's acceptance, while \methodname{} remains faster than DFlash.}
Accepted length $\tau$ (solid) and throughput (dashed), relative
to DFlash; Shading highlights the separation between the accepted-length and throughput ratios..
Qwen3-4B, thinking off, $T=0$ (dark) and $T=1$ (light),
averaged over all ten benchmark suites. Absolute values:
Tab.~\ref{tab:concurrency_q4b}.}
    \label{fig:q4b_concurrency}
\end{figure}

\begin{figure}[t]
    \centering
    \includegraphics[width=\linewidth]{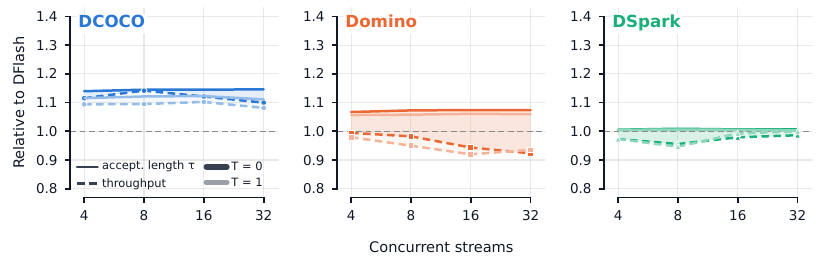}

    \caption{\textbf{At high concurrency, overhead outweighs Domino's acceptance gain and DSpark gains no acceptance over DFlash, while \methodname{} remains faster than DFlash.}
Accepted length $\tau$ (solid) and throughput (dashed), relative
to DFlash; Shading highlights the separation between the accepted-length and throughput ratios.
Qwen3-Coder-30B-A3B, thinking off, $T=0$ (dark) and $T=1$ (light),
averaged over all ten benchmark suites. Absolute values:
Tab.~\ref{tab:concurrency_q30b}.}
    \label{fig:q30b_concurrency}
\end{figure}

\begin{figure}[t]
    \centering
    \includegraphics[width=\linewidth]{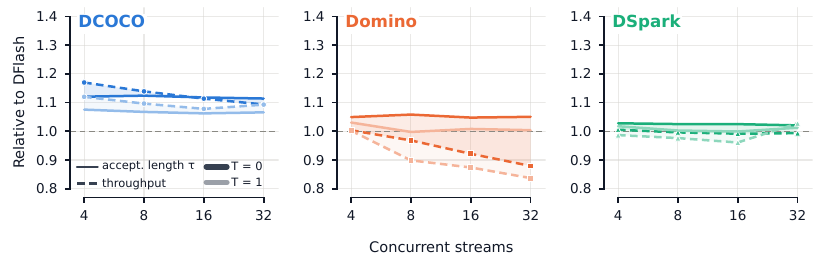}

    \caption{\textbf{As concurrency grows, Domino's acceptance gain is consumed by per-cycle overhead, while \methodname{} remains faster than DFlash.}
Accepted length $\tau$ (solid) and throughput (dashed), relative
to DFlash; Shading highlights the separation between the accepted-length and throughput ratios.
Llama-3.1-8B, thinking off, $T=0$ (dark) and $T=1$ (light),
averaged over all ten benchmark suites. Absolute values:
Tab.~\ref{tab:concurrency_llama}.}
    \label{fig:llama_concurrency}
\end{figure}

Batched serving is where correction overhead either pays or does not:
as concurrency grows, the verifier's cycle time is amortised across
streams, so whatever a corrector adds per cycle weighs ever more
heavily against the acceptance it buys. This appendix extends the
main-text view of Figures~\ref{fig:q8b_conc}
and~\ref{fig:concurrency} to the remaining backbones.
Figures~\ref{fig:q4b_conc}--\ref{fig:llama_conc} plot aggregate
throughput with \methodname's margin over the strongest baseline at
each load, and Tables~\ref{tab:concurrency}--%
\ref{tab:concurrency_llama} give the absolute numbers from
concurrency~1: \methodname{} is the fastest arm in every cell, all
four backbones, both temperatures, by $6$--$23\%$ over the
strongest baseline, while Domino drops below even the uncorrected
drafter as load grows and DSpark sits at or below it throughout.
Figures~\ref{fig:q4b_concurrency}--\ref{fig:llama_concurrency} show why,
plotting each corrector's accepted length and throughput relative to
DFlash with its per-cycle overhead as the shaded gap between the two
curves: Domino's acceptance gain persists at every load, but the
widening overhead consumes it, the $\tau$ curve stays above parity
while the throughput curve slides below, DSpark buys little or no
acceptance to begin with, and \methodname's two curves move together
above parity because its shading stays thin.

\begin{table}[h]
\centering
\small
\begin{tabular}{rcccccc}
\toprule
Concurrency & DFlash & Domino & DSpark & \methodname & \methodname / DFlash & \methodname vs.\ best \\
\midrule
\multicolumn{7}{l}{\emph{Greedy decoding, $T{=}0$}} \\
1 & 783.1 & 839.3 & 765.6 & \best{959.7} & $1.23\times$ & $+14.3\%$ \\
4 & 2806.4 & 3008.6 & 2719.0 & \best{3507.3} & $1.25\times$ & $+16.6\%$ \\
8 & 4889.1 & 5051.0 & 4614.4 & \best{6082.6} & $1.24\times$ & $+20.4\%$ \\
16 & 7229.1 & 7098.4 & 7092.9 & \best{8894.3} & $1.23\times$ & $+23.0\%$ \\
32 & 9655.0 & 9013.9 & 9490.3 & \best{11512.4} & $1.19\times$ & $+19.2\%$ \\
\midrule
\multicolumn{7}{l}{\emph{Sampling, $T{=}1$ (mean over three seeds)}} \\
1 & 649.8 & 670.9 & 628.7 & \best{762.2} & $1.17\times$ & $+13.6\%$ \\
4 & 2300.3 & 2334.1 & 2198.9 & \best{2710.0} & $1.18\times$ & $+16.1\%$ \\
8 & 4006.3 & 3894.9 & 3758.1 & \best{4735.3} & $1.18\times$ & $+18.2\%$ \\
16 & 5964.5 & 5441.1 & 5730.3 & \best{6871.9} & $1.15\times$ & $+15.2\%$ \\
32 & 7857.9 & 6821.8 & 7443.4 & \best{8836.1} & $1.12\times$ & $+12.4\%$ \\
\bottomrule
\end{tabular}
\caption{\textbf{Aggregate throughput (output tokens/s) as concurrency grows}, Qwen3-8B, thinking
off, averaged over the ten suites per concurrency workload.}
\label{tab:concurrency}
\end{table}

\begin{table}[h]
\centering
\small
\begin{tabular}{ccccccc}
\toprule
Concurrency & DFlash & Domino & DSpark & \methodname & \methodname / DFlash & \methodname vs.\ best \\
\midrule
\addlinespace[3pt]
\multicolumn{7}{l}{\emph{Greedy decoding, $T{=}0$}} \\
\addlinespace[3pt]
1 & 1096.9 & 1202.7 & 980.1 & \best{1332.6} & $1.21\times$ & $+10.8\%$ \\
4 & 4097.1 & 4032.1 & 3517.5 & \best{4660.3} & $1.14\times$ & $+13.7\%$ \\
8 & 6942.6 & 6560.9 & 5953.9 & \best{7781.7} & $1.12\times$ & $+12.1\%$ \\
16 & 10940.3 & 9633.9 & 9448.6 & \best{12215.5} & $1.12\times$ & $+11.7\%$ \\
32 & 14181.1 & 11966.5 & 12746.7 & \best{15602.7} & $1.10\times$ & $+10.0\%$ \\
\midrule
\addlinespace[3pt]
\multicolumn{7}{l}{\emph{Sampling, $T{=}1$ (mean over three seeds)}} \\
\addlinespace[3pt]
1 & 927.6 & 995.7 & 838.4 & \best{1086.8} & $1.17\times$ & $+9.1\%$ \\
4 & 3345.6 & 3226.4 & 2929.6 & \best{3721.2} & $1.11\times$ & $+11.2\%$ \\
8 & 5697.0 & 5306.7 & 5015.9 & \best{6191.9} & $1.09\times$ & $+8.7\%$ \\
16 & 9092.8 & 7830.4 & 7965.6 & \best{9834.0} & $1.08\times$ & $+8.2\%$ \\
32 & 11676.9 & 9628.3 & 10525.8 & \best{12476.2} & $1.07\times$ & $+6.8\%$ \\
\bottomrule
\end{tabular}
\caption{\textbf{Aggregate throughput (output tokens/s) as concurrency grows}, Qwen3-4B, thinking off, averaged over the ten suites. Bold: best method per row.}
\label{tab:concurrency_q4b}
\end{table}

\begin{table}[h]
\centering
\small
\begin{tabular}{ccccccc}
\toprule
Concurrency & DFlash & Domino & DSpark & \methodname & \methodname / DFlash & \methodname vs.\ best \\
\midrule
\addlinespace[3pt]
\multicolumn{7}{l}{\emph{Greedy decoding, $T{=}0$}} \\
\addlinespace[3pt]
1 & 642.4 & 660.1 & 627.2 & \best{785.1} & $1.22\times$ & $+18.9\%$ \\
4 & 1862.6 & 1852.5 & 1812.2 & \best{2077.2} & $1.12\times$ & $+11.5\%$ \\
8 & 3110.6 & 3055.5 & 2972.4 & \best{3550.5} & $1.14\times$ & $+14.1\%$ \\
16 & 4886.7 & 4610.1 & 4784.7 & \best{5477.8} & $1.12\times$ & $+12.1\%$ \\
32 & 7155.2 & 6606.8 & 7057.2 & \best{7867.9} & $1.10\times$ & $+10.0\%$ \\
\midrule
\addlinespace[3pt]
\multicolumn{7}{l}{\emph{Sampling, $T{=}1$ (mean over three seeds)}} \\
\addlinespace[3pt]
1 & 556.9 & 557.8 & 540.0 & \best{662.3} & $1.19\times$ & $+18.7\%$ \\
4 & 1601.0 & 1567.8 & 1557.3 & \best{1751.2} & $1.09\times$ & $+9.4\%$ \\
8 & 2679.3 & 2544.9 & 2535.8 & \best{2933.2} & $1.09\times$ & $+9.5\%$ \\
16 & 4185.4 & 3849.4 & 4154.3 & \best{4613.0} & $1.10\times$ & $+10.2\%$ \\
32 & 5892.4 & 5511.0 & 5890.4 & \best{6371.9} & $1.08\times$ & $+8.1\%$ \\
\bottomrule
\end{tabular}
\caption{\textbf{Aggregate throughput (output tokens/s) as concurrency grows}, Qwen3-Coder-30B-A3B, thinking off, averaged over the ten suites. Bold: best method per row.}
\label{tab:concurrency_q30b}
\end{table}

\begin{table}[h]
\centering
\small
\begin{tabular}{ccccccc}
\toprule
Concurrency & DFlash & Domino & DSpark & \methodname & \methodname / DFlash & \methodname vs.\ best \\
\midrule
\addlinespace[3pt]
\multicolumn{7}{l}{\emph{Greedy decoding, $T{=}0$}} \\
\addlinespace[3pt]
1 & 844.2 & 900.8 & 838.0 & \best{1042.4} & $1.23\times$ & $+15.7\%$ \\
4 & 3128.0 & 3139.2 & 3145.0 & \best{3659.2} & $1.17\times$ & $+16.3\%$ \\
8 & 5388.7 & 5217.4 & 5367.7 & \best{6137.4} & $1.14\times$ & $+13.9\%$ \\
16 & 8007.2 & 7379.9 & 7940.5 & \best{8916.8} & $1.11\times$ & $+11.4\%$ \\
32 & 10730.3 & 9435.8 & 10660.3 & \best{11731.7} & $1.09\times$ & $+9.3\%$ \\
\midrule
\addlinespace[3pt]
\multicolumn{7}{l}{\emph{Sampling, $T{=}1$ (mean over three seeds)}} \\
\addlinespace[3pt]
1 & 487.2 & 497.3 & 459.6 & \best{536.3} & $1.10\times$ & $+7.8\%$ \\
4 & 1724.9 & 1728.4 & 1702.1 & \best{1931.8} & $1.12\times$ & $+11.8\%$ \\
8 & 2969.0 & 2668.5 & 2898.7 & \best{3255.2} & $1.10\times$ & $+9.6\%$ \\
16 & 4308.5 & 3765.6 & 4140.3 & \best{4644.4} & $1.08\times$ & $+7.8\%$ \\
32 & 5396.9 & 4517.5 & 5547.6 & \best{5897.7} & $1.09\times$ & $+6.3\%$ \\
\bottomrule
\end{tabular}
\caption{\textbf{Aggregate throughput (output tokens/s) as concurrency grows}, Llama-3.1-8B, thinking off, averaged over the ten suites. Bold: best method per row.}
\label{tab:concurrency_llama}
\end{table}

\section{Stochastic Verification with Deterministic Proposals}
\label{app:deterministic-proposals}

\methodname selects draft tokens deterministically using the prefix-first rule,
including when the verifier uses stochastic decoding. Conditional on the
current context and the selected draft prefix, its proposal distribution is
therefore a point mass:
\begin{equation}
    p_j(v \mid x,\hat y_{<j}) = \mathbf{1}[v=\hat y_j].
    \label{eq:dcoco-point-mass}
\end{equation}
The probability one refers to proposal selection, not verifier acceptance.
Let $q_j(v)=q(v\mid x,\hat y_{<j})$ denote the verifier distribution after
applying the target sampling settings, including temperature. Standard
speculative sampling accepts $\hat y_j$ with probability
$\min\{1,q_j(\hat y_j)/p_j(\hat y_j)\}=q_j(\hat y_j)$.
On rejection, its residual distribution simplifies to
\begin{equation}
    r_j(v)=\frac{q_j(v)\,\mathbf{1}[v\ne\hat y_j]}
                    {1-q_j(\hat y_j)}.
    \label{eq:dcoco-point-mass-residual}
\end{equation}
When $q_j(\hat y_j)=1$, rejection never occurs and the residual is not needed.

This rule preserves the verifier distribution: the proposed token is emitted
with probability $q_j(\hat y_j)$, while any other token $v$ is emitted with
probability $(1-q_j(\hat y_j))r_j(v)=q_j(v)$. Verification continues through
accepted draft tokens, commits the residual sample at the first rejection,
and discards the remaining draft suffix. If all draft tokens are accepted,
one additional token is sampled from the verifier. Applying the same argument
at each committed position preserves the target autoregressive distribution.

Consequently, stochastic target decoding does not require stochastic draft
selection: \methodname's transition scores need not be softmax-normalized, and no
distribution over the retained candidates needs to be sampled or passed to
the verifier. The verifier must use the point-mass proposal semantics above.

\end{document}